\documentclass[11pt]{article}

\usepackage[preprint]{acl}

\usepackage{times}
\usepackage{latexsym}

\usepackage[T1]{fontenc}

\usepackage[utf8]{inputenc}

\usepackage{microtype}

\usepackage{inconsolata}

\usepackage{graphicx}

\usepackage{times}
\usepackage{latexsym}

\usepackage[T1]{fontenc}

\usepackage[utf8]{inputenc}

\usepackage{microtype}

\usepackage{inconsolata}

\usepackage{tabularx}
\usepackage{multirow}
\usepackage{soul}
\usepackage{graphicx}
\usepackage{booktabs}
\usepackage{rotating}
\usepackage[table]{xcolor}
\usepackage{color, colortbl}
\usepackage{changepage,threeparttable}
\usepackage[htt]{hyphenat}
\usepackage{hyperref}
\usepackage{afterpage}
\usepackage{float}
\usepackage{changepage}
\usepackage{setspace}

\usepackage[most]{tcolorbox}
\usepackage{xcolor}

\newtcolorbox{promptbox}[1][]{
    colback=gray!5,
    colframe=gray!75,
    fonttitle=\bfseries,
    left=5pt,
    right=5pt,
    top=5pt,
    bottom=5pt,
    breakable,
    fontupper=\small\ttfamily,
    fontlower=\small\ttfamily,
    #1
}

\newcommand{\promptcomment}[1]{\textcolor{red}{\textit{#1}}}

\newtcolorbox{inputbox}[1][]{
    colback=pink!20,
    colframe=pink!75,
    fonttitle=\bfseries,
    left=5pt,
    right=5pt,
    top=5pt,
    bottom=5pt,
    breakable,
    fontupper=\small\ttfamily,
    fontlower=\small\ttfamily,
    #1
}

\newtcolorbox{responsebox}[1][]{
    colback=gray!5,
    colframe=gray!75,
    fonttitle=\bfseries,
    left=5pt,
    right=5pt,
    top=5pt,
    bottom=5pt,
    breakable,
    fontupper=\small\ttfamily,
    fontlower=\small\ttfamily,
    #1
}

\newcommand{\errorcomment}[1]{\textcolor{red}{\textit{#1}}}

\usepackage{listings}
\usepackage{xcolor}

\title{Exploring State Tracking Capabilities of Large Language Models}

\author{Kiamehr Rezaee ,  Jose Camacho-Collados \and Mohammad Taher Pilehvar \\ Cardiff NLP, School of Computer Science and Informatics \\
\{rezaeek, camachocolladosj, pilehvarmt\}@cardiff.ac.uk}

\begin{document}
\maketitle

\newcommand{\kia}[1]{\textcolor{orange}{#1}}
\newcommand{\td}[1][]{\textcolor{red}{TODO #1}}
\newcommand{\jose}[1]{\textcolor{blue}{#1}}

\begin{spacing}{0.98}
\begin{abstract}

Large Language Models (LLMs) have demonstrated impressive capabilities in solving complex tasks, including those requiring a certain level of reasoning. In this paper, we focus on state tracking, a problem where models need to keep track of the state governing a number of entities. To isolate the state tracking component from other factors, we propose a benchmark based on three well-defined state tracking tasks and analyse the performance of LLMs in different scenarios. The results indicate that the recent generation of LLMs (specifically, GPT-4 and Llama3) are capable of tracking state, especially when integrated with mechanisms such as Chain of Thought. 
However, models from the former generation, while understanding the task and being able to solve it at the initial stages, often fail at this task after a certain number of steps.

\end{abstract}

\section{Introduction}
State tracking refers to the process of maintaining and updating an internal representation of a system's condition or status over time, based on a sequence of observations or inputs. This internal representation, or the ``state'', encapsulates relevant information from past inputs that is necessary to predict future outputs or behaviors of the system. 
%
This capability comprises three essential components: an internal representation or model of the system or ``world'', the capacity to comprehend the effects of actions and updates on this world, and some form of on-demand memory system that maintains the current and updated versions of the world state.


This phenomenon has been studied in Large Language Models (LLMs) based on the Transformer architecture \cite{vaswani2017attention}, highlighting potential shortcomings of these systems in tracking the state of games such as Othello and chess  \cite{li2023emergent, Toshniwal2021ChessAA}. The possible reasons behind these flaws could be attributed to two main factors: 
(1) the single-modality text-based pre-training objective, which may not provide a comprehensive model of the real world for the system or may offer insufficient training signals, thereby limiting the model's ability to effectively integrate and utilize contextual information, and (2) inherent issues with the underlying Transformer architecture, which might be inadequate for tasks that require consistent state tracking.


In this context, theoretical investigations into the computational boundaries of the Transformer architecture have revealed limitations, particularly regarding their ability to express sequential computation and the challenge of tracking a coherent state through intermediate steps while solving a problem \cite{merrill-sabharwal-2023-parallelism, dehghani2018universal}. However,  \newcite{li2024chain} suggested that in theory, by incorporating Chain of Thought, transformers can overcome their limitations and solve problems that are inherently serial and not parallelized (e.g. most state tracking tasks). One shortcoming of these theoretical studies is that they have not examined such limitations in the context of LLMs trained on natural language, leaving unanswered the open question of whether LLMs can acquire the capability of state tracking through training on general natural language data.


The main research question in this paper is whether state-of-the-art Transformer-based LLMs, trained on natural language textual data, can effectively manage state tracking.
To this end, we provide a comprehensive assessment of the ability of LLMs to track states by designing a set of probing tasks. These tasks were designed 
to offer deeper insights into the capabilities and limitations of LLMs in maintaining consistent states throughout interactions. To limit confound effects, we have presented the tasks in a simple form along with baselines showing that LLMs are able to understand the tasks, so as to  to isolate the effect of state tracking capabilities. We diversified the datasets by incorporating various configurations and designed the tasks to be as simple as possible for human solvers. 

Our main contributions are threefold: (1) We propose a simple process to construct datasets for three carefully designed elementary tasks to highlight LLMs' competence in state tracking, which can be used both as a benchmark and as a sanity check when deploying LLMs\footnote{We will release the simulation code used to generate the task datasets, along with the datasets used in our experiments upon acceptance.}; (2) We perform a set of detailed experiments across LLMs to measure their performance in state tracking, isolating it from other factors; and (3) We provide an in-depth analysis to shed light on the possible factors impacting the performance of various systems.

    
\section{Related Work}
\subsection{Theoretical Perspective}
\label{sec:state_tracking}
From an algebraic formal language theory perspective, previous studies have laid the theoretical foundation by forming connection between the problem of state tracking and the well-studied mathematical field of group theory \cite{liu2023transformers, Merrill2024TheIO}. 
This connection provides a way to establish the computational complexity class in which different state tracking problems reside, and a theoretical upper-bound to the complexity class of problems that are solvable by Transformer architecture has been provided \cite{merrill-sabharwal-2023-parallelism}.
These studies suggest that tasks that are inherently ``non-parallelizable'' in nature and involve some sort of recurrence mechanism in their solution (e.g. most of the state tracking problems with the need for memory-based storage of intermediate states during repeated update applications), are ``hard'' for the Transformer architecture to solve.


The work by \cite{li2024chain} further extends these theoretical studies by suggesting that Transformer based models, when allowed to generate as many intermediate reasoning tokens as needed, can utilize their input window as a form of memory, effectively making them powerful enough to solve such inherently serial tasks.


A key distinction of these works to ours is their focus on transformers trained with specially designed objectives tailored to specific tasks. In contrast, we explore transformers in the context of LLMs trained using the language modeling objective, and investigate whether these models naturally learn to utilize their input tape as a form of intermediate memory to track reasoning steps and states, when trained on natural language data.

 


\subsection{Memory Integration}

An essential ability required to solve state tracking problem is to have an internal memory capable of storing and updating internal states. A fundamental drawback of the models based on the Transformer architecture compared to naturally recurrent architectures like Recurrent Neural Networks (e.g. RNNs), is that the only form of inference-time memory accessible to Transformers is their limited input window, which they can write to, whereas in the case of RNNs, in theory, they can update their internal representation of state infinite times.

According to \citet{dehghani2018universal}, the Transformer architecture, can become Turing-complete (hence able to solve state tracking problems) simply by integrating some form of read-write memory.

The evidence of improvements when integrating memory with Transformers, has been reported in various studies \cite{zhong-etal-2022-training, Wu2022MemorizingT}. This proposed memory components are mainly suggested as a way to improve performance on tasks that rely on retrieving relevant information from documents. However, this type of memory augmentation might not be as advantageous for state tracking problems as they are not necessarily well-equipped to effectively utilize it for problem-solving tasks that involve tracking and updating states sequentially in multiple steps.

Perhaps more inline with theoretical findings of \citet{li2024chain}, in the context of unmodified Transformer-based LLMs, \citet{wei2022chain} propose a method, namely Chain of Though (CoT), in which the language model is simply instructed to explain the intermediate steps of solving the problem at hand, effectively guiding the model to use its input window as some sort of temporary memory.

In this work, as we are interested to see whether the Transformer-based models, with no further modifications, are able to handle state tracking problems by utilizing their limited input tape, we will provide evaluations of systems prompted in this manner, although other type of memory integration could be tried.


\subsection{Multi-step Reasoning and State Tracking Evaluation} \label{related_work:tasks}


Previous studies have aimed to introduce relevant probing tasks in order to assess the proficiency of LLMs in tracking the underlying state that governs entities and their interactions throughout conversations. 
In this context, a branch of studies explore the capabilities of language models in dialogue state tracking \cite{williams-etal-2013-dialog, budzianowski-etal-2018-multiwoz}. These studies typically assess how well systems respond to queries based on the context provided in the chat history. Our work, however, differs in a key aspect: we investigate whether models can answer queries about state and information not explicitly present in the dialogue history.

Another branch of studies focuses on the ability of models to perform multi-hop reasoning, that involves multiple intermediate steps of inference \cite{yang-etal-2018-hotpotqa, welbl2018constructing, khot2020qasc}. These studies differ from ours as they assess models in terms of the ability to integrate information from diverse contexts and sources, which involves various forms of reasoning, such as logical and commonsense reasoning. In contrast, our work primarily involves procedural and sequential reasoning, where the task is to accurately apply a series of update statement changes. This allows for a detailed analysis of the model's capability by introducing a controlled experimental environment.

Perhaps more aligned with the state tracking capability we are interested in is the \texttt{chess\_state\_tracking}\footnote{The dataset can be found at \url{https://github.com/google/BIG-bench/tree/main/bigbench/benchmark_tasks/chess_state_tracking}} task in the Big-bench benchmark suite \cite{srivastava2023imitation} could be mentioned. As the names suggest, the task is querying model about the final state of a chess board after a series of updates. However, this task does not fully capture what we are trying to measure, as the update statements are presented in an algebraic chess notation, which abstracts away the nuances of natural language.

Moreover,\texttt{tracking\_shuffled\_objects}\footnote{Dataset at \url{https://github.com/google/BIG-bench/tree/main/bigbench/benchmark_tasks/tracking_shuffled_objects}} task in Big-bench describes a set of objects that are held by a number of individuals and are swapped in a series of pairwise trades, and the model is asked about the final placement of an object. However, a notable challenge arises from the number of the update statements presented in this task, which lacks the necessary diversity to support comprehensive analysis. 

\section{Task Design}

In this section, we outline the design perspective we applied throughout the process of creating the set of simple tasks. Our goals are two-fold: (1) first, to keep the tasks simple enough for humans to solve, and (2) second, to provide different scenarios where the ability to track some form of internal state is essential to solving the given task. Aggregating the results of different tasks with various wordings also helps to isolate the state tracking aspects we are interested in and eliminate the variance in performance caused by several factors such as the surface form (the wording) of the prompts designed for each task. 

\begin{figure}
    \centering
    \includegraphics[width=\linewidth]{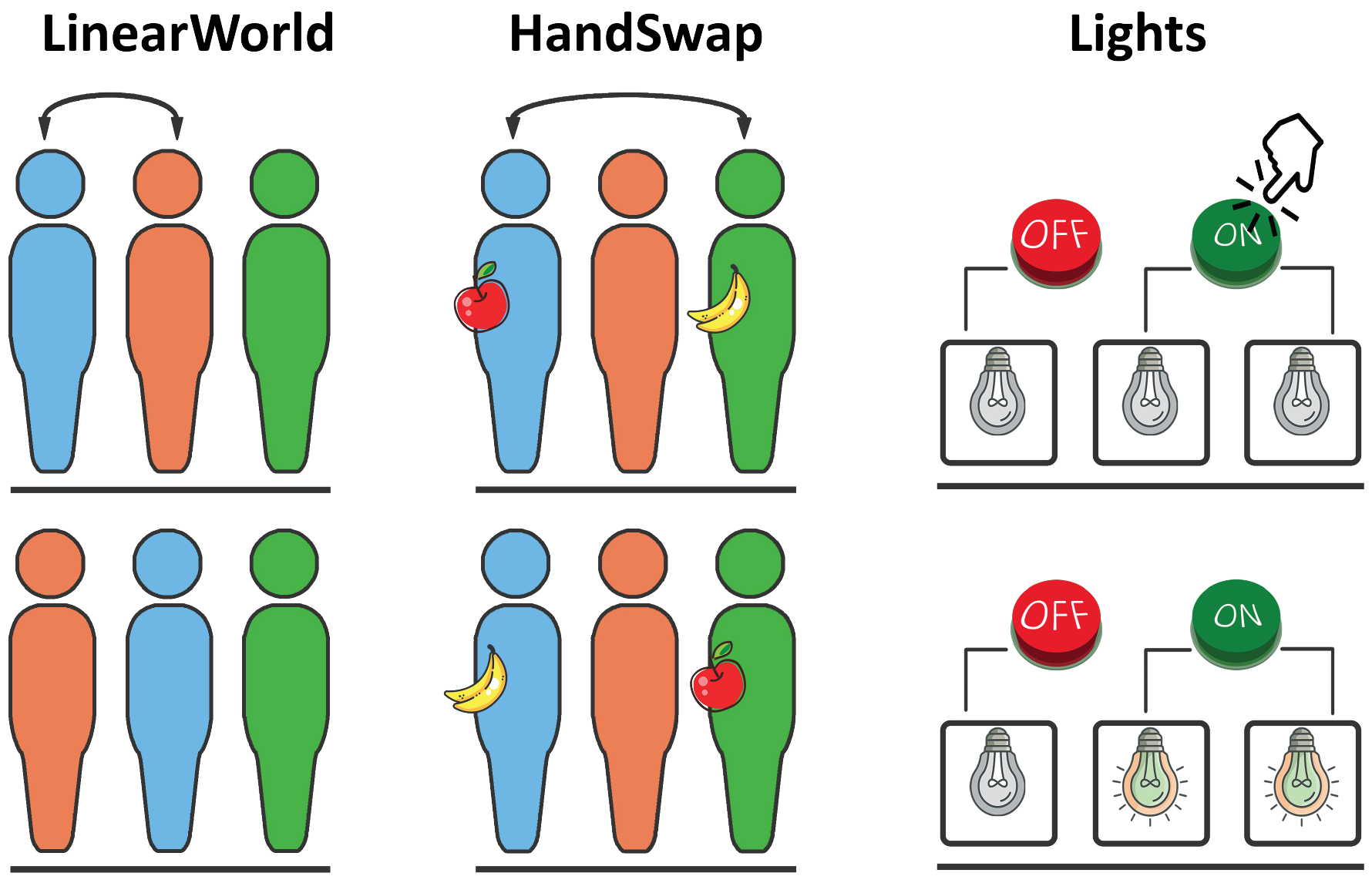}
    \caption{A simple illustration of the initial state of the tasks (top) and the updated state after a singular update step (bottom) of the tasks. Note that the configuration shown for \texttt{Lights} is not complete as only two switches are depicted in the image; for the exact configuration of rooms and switches, please refer to section \ref{sec:task_configuration}.}
    \label{fig:tasks}
\end{figure}

Each task in this set consists of an initial state of an auxiliary world (containing entities and/or objects) presented to the model, followed by a series of elementary updates to the aforementioned world, and finally a query about the final state of the world. We use the term "depth" to refer to the number of world updates provided to the model before presenting the query. In other words, the depth indicates the recurrence required for the model to apply the transition function from the initial state to obtain the final state.

We present three different tasks, referred to as \textbf{\texttt{LinearWorld}}, \textbf{\texttt{HandSwap}}, and \textbf{\texttt{Lights}}. The first two are different representations of the same underlying problem, which involves tracking different permutation configurations of a number of objects. An illustration of these three tasks is presented at Figure \ref{fig:tasks}. We explain each task separately in the following subsections.

\subsection{LinearWorld}
The first task in this suite of tasks consists of an auxiliary linear world where a number of entities are standing on a line in a specific order. The initial configuration of this world is presented to the model by the name and position of entities, with an integer indicating their position (e.g. "entity A is standing at point 3") relative to the origin at coordinate (0). The state of the world changes whenever two entities swap their positions (e.g., 'Entity A swaps positions with Entity B'). Finally, the query in this setting, is simply a choice from list of possible relative position of two arbitrary entities (e.g. "Where is entity A standing with respect to entity C now?" with options being ["right", "left"]). An example of the prompt we use for this task in provided in Appendix~\ref{app:prompt_examples:linear}.

\subsection{HandSwap}
Another form of the same underlying problem as the \texttt{LinearWorld} task (which deals with permutation combinations of a number of entities) is where the update statements address indices rather than the entities themselves. For example, in case of the \texttt{LinearWorld} task, an update might be given as "whoever is standing in position 3 changes positions with whoever is standing in position 1," rather than "entity A swaps positions with entity B." This task is an extension to the \textit{tracking\_shuffled\_objects} task in the Big-bench suite \cite{srivastava2023imitation}, in which we present more depth (number of update statements) to the instances. 

As for this task, the auxiliary world consist of a number of individuals holding various items in their hands.  The initial configuration of this world is described by the name of entities, along with the name of the item they are holding at the moment. Each update involves two individuals in the world exchanging the item in their hands ("e.g. "entity A swaps hands with entity B"). The query involves selecting an item from a list of potential objects that an arbitrary individual of choice might be holding (e.g. "What is entity A holding in their hands right now?" with options such as ["apple","banana", "nothing"]). An example of the prompt we use for this task in provided in Appendix~\ref{app:prompt_examples:hands}.

Note that here, the subjects of the updates are the holdable items, while the entities act as the containers or indices. Although the underlying problem remains the same as the \texttt{LinearWorld} problem, by introducing variety through different wordings compared to the \texttt{LinearWorld} task, we aim to minimize the impact of surface form bias in our evaluations.

\subsection{Lights}
The final task involves a fundamentally different problem compared to the other two. In this task, the model of world contains a number of rooms each having a light bulb in them. The auxiliary world also contains a number of individual switches. The initial configuration of this world is defined by the status of each light in each room (e.g. "The light in room A is initially off"), and how the functionality of each switch affects the light in each room (e.g. "When switch 1 is pressed, it turns the light of room A off"). The updates here are simply represented by a series of key presses (e.g. "The switch 4 is pressed"), and the query in this configuration asks for the final status of the light in a room of choice (e.g. "What is the final light status in room C?" with options being ["off", "on"]). An example of the prompt we use for this task in provided in Appendix \ref{app:prompt_examples:lights}.

\begin{figure*}[h!]

        \includegraphics[width=\linewidth]{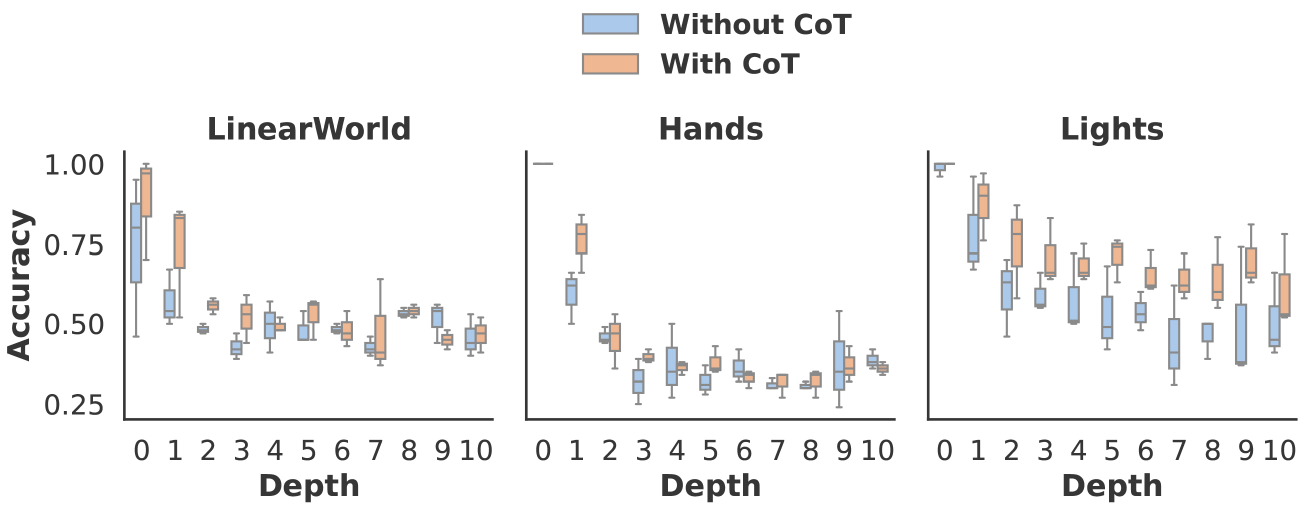}
        
    \caption{Average accuracy at different depths across tasks (left: \texttt{LinearWorld}, middle: \texttt{Hands}, and right: \texttt{Lights}) for all systems except for the two top performers which use Chain of Thought (CoT), i.e., Llama3 70B and GPT-4.}
    \label{fig:cot_impact}

\end{figure*}

\section{Experimental Setting}
In this section, we describe the various experimental configurations used throughout our experiments.

\subsection{Comparison Systems}

Our main group of evaluation systems consists of Transformer-based LLMs, specifically GPT-3.5-turbo, GPT-4o, Mixtral-8x7B-Instruct \cite{jiang2024mixtralexperts}, LLama3-8B and Llama3-70B\footnote{At the time this paper was written, there were no official papers available for GPT-4o or Llama3 models. Please refer to the respective blog posts at the following addresses: GPT-4o: \url{https://openai.com/index/hello-gpt-4o/}, Llama3: \url{https://llama.meta.com/llama3/}.}. We evaluate these models via a prompt-based approach. Each prompt consists of an instruction statement that sets the model up for the task at hand, a natural language representation of the initial state, a series of update statements, and finally, a query presented in natural language form.


In addition, to enhance model performance, we provide a solved example to the model in advance. Moreover, since the input tape of transformers can function as a form of on-demand memory for solving tasks, we conducted a variation of experiments with Chain of Thought (CoT) instructed models. Specifically, we instructed the models to include their reasoning process in the output to assess how impactful this type of memory is for task solving. Example of the prompts used for each task can be found in Appendix~\ref{app:prompt_examples}.

\subsection{Task Configurations}
\label{sec:task_configuration}

Regarding the number of entities in the main configuration of the \textbf{\texttt{LinearWorld}} task, we opt for three individuals in this world to simplify the task for Transformers. However, we also conducted complementary experiments with various configurations of this task, including experiments with five entities, as discussed in section \ref{sec:complementary_study}. As for \textbf{\texttt{HandSwap}} task, similar to \texttt{LinearWorld}, the world consist of three individuals. Moreover, for the number of objects, we opt for three holdable options: "apple," "banana," and "nothing," which are swapped between the three entities. In the \textbf{\texttt{Lights}} task, we opt for three rooms and five switches. Each of the switches has different functionalities, as illustrated in the example prompt in Appendix~\ref{app:prompt_examples:lights}. These numbers are selected to ensure a manageable yet challenging range of possible states.

For each task, we developed a simulator to model the target world. Starting from an initial state, we sampled 50 episodes for each experiment, where each episode represents a trajectory from the initial state presented to the model to the final state after a sequence of 10 update statements, along with the corresponding intermediate states that occurred during these updates. The model is presented to these episodes in a prompt template format. The generation temperature parameter is set to ensure that the model's responses are deterministic. Additionally, we apply minimal regex processing to the model's output to extract the final label. To evaluate performance, we computed the accuracy of the models' answers to queries at each depth of the update sequences across all the episodes.

\section{Results}
Table \ref{tab:main_results} illustrates the results of the evaluations. In what follows, we discuss some of the observations we have made:

    




\begin{table*}[!ht]
    \resizebox{\textwidth}{!}{\begin{tabular}{ll ccccccccccc}\toprule
\multirow{2}{*}{Task} &\multirow{2}{*}{System} &\multicolumn{11}{c}{Depth} \\\cmidrule(lr){3-13}
& &0 &1 &2 &3 &4 &5 &6 &7 &8 &9 &10 \\

\midrule

\multirow{12}{*}{Linear} &GPT 3.5 &\cellcolor[HTML]{6ac397}0.95 &0.54 &\cellcolor[HTML]{f7d7d4}0.48 &\cellcolor[HTML]{ec9c95}0.39 &\cellcolor[HTML]{eea9a3}0.41 &\cellcolor[HTML]{f3c4bf}0.45 &\cellcolor[HTML]{f9e4e2}0.50 &\cellcolor[HTML]{eda39d}0.40 &\cellcolor[HTML]{fdf8f8}0.53 &0.54 &\cellcolor[HTML]{fdf8f8}0.53 \\
&GPT 3.5 CoT &\cellcolor[HTML]{62c092}0.97 &\cellcolor[HTML]{8ed2b1}0.85 &\cellcolor[HTML]{f8fdfa}0.56 &\cellcolor[HTML]{edf8f3}0.59 &\cellcolor[HTML]{f7d7d4}0.48 &\cellcolor[HTML]{f3c4bf}0.45 &\cellcolor[HTML]{f6d1cd}0.47 &\cellcolor[HTML]{e98f87}0.37 &\cellcolor[HTML]{f8fdfa}0.56 &\cellcolor[HTML]{f3c4bf}0.45 &\cellcolor[HTML]{f6d1cd}0.47 \\
&Mixtral &\cellcolor[HTML]{a1d9bd}0.80 &\cellcolor[HTML]{d0ecde}0.67 &\cellcolor[HTML]{f6d1cd}0.47 &\cellcolor[HTML]{f6d1cd}0.47 &\cellcolor[HTML]{f5fbf8}0.57 &\cellcolor[HTML]{f3c4bf}0.45 &\cellcolor[HTML]{f6d1cd}0.47 &\cellcolor[HTML]{efb0aa}0.42 &\cellcolor[HTML]{fcfefd}0.55 &\cellcolor[HTML]{f2bdb8}0.44 &\cellcolor[HTML]{eda39d}0.40 \\
&Mixtral CoT &\cellcolor[HTML]{57bb8a}1.00 &\cellcolor[HTML]{96d5b6}0.83 &\cellcolor[HTML]{fdf8f8}0.53 &\cellcolor[HTML]{fdf8f8}0.53 &\cellcolor[HTML]{f7d7d4}0.48 &\cellcolor[HTML]{f5fbf8}0.57 &\cellcolor[HTML]{f1b6b1}0.43 &\cellcolor[HTML]{eea9a3}0.41 &0.54 &\cellcolor[HTML]{f7d7d4}0.48 &\cellcolor[HTML]{eea9a3}0.41 \\
&Llama3 8B &\cellcolor[HTML]{f5cac7}0.46 &\cellcolor[HTML]{f9e4e2}0.50 &\cellcolor[HTML]{f9e4e2}0.50 &\cellcolor[HTML]{efb0aa}0.42 &\cellcolor[HTML]{f9e4e2}0.50 &0.54 &\cellcolor[HTML]{f7d7d4}0.48 &\cellcolor[HTML]{f5cac7}0.46 &\cellcolor[HTML]{fcf1f0}0.52 &\cellcolor[HTML]{f8fdfa}0.56 &\cellcolor[HTML]{f2bdb8}0.44 \\
&Llama3 8B - CoT &\cellcolor[HTML]{c5e8d7}0.70 &\cellcolor[HTML]{fcf1f0}0.52 &\cellcolor[HTML]{f1faf5}0.58 &\cellcolor[HTML]{f2bdb8}0.44 &\cellcolor[HTML]{fcf1f0}0.52 &\cellcolor[HTML]{f8fdfa}0.56 &0.54 &\cellcolor[HTML]{dbf1e6}0.64 &\cellcolor[HTML]{fcf1f0}0.52 &\cellcolor[HTML]{efb0aa}0.42 &\cellcolor[HTML]{fcf1f0}0.52 \\
&Llama3 70B &\cellcolor[HTML]{57bb8a}1.00 &\cellcolor[HTML]{99d6b8}0.82 &0.54 &\cellcolor[HTML]{e67c73}0.34 &\cellcolor[HTML]{f8fdfa}0.56 &\cellcolor[HTML]{d4eee1}0.66 &\cellcolor[HTML]{fcf1f0}0.52 &\cellcolor[HTML]{fcf1f0}0.52 &\cellcolor[HTML]{e2f4eb}0.62 &\cellcolor[HTML]{f8fdfa}0.56 &\cellcolor[HTML]{fcf1f0}0.52 \\
&Llama3 70B - CoT &\cellcolor[HTML]{66c195}0.96 &\cellcolor[HTML]{8bd0ae}0.86 &\cellcolor[HTML]{83cda9}0.88 &\cellcolor[HTML]{a1d9bd}0.80 &\cellcolor[HTML]{a8dcc2}0.78 &\cellcolor[HTML]{7ccaa4}0.90 &\cellcolor[HTML]{99d6b8}0.82 &\cellcolor[HTML]{8bd0ae}0.86 &\cellcolor[HTML]{afdfc8}0.76 &\cellcolor[HTML]{bee5d2}0.72 &\cellcolor[HTML]{92d3b3}0.84 \\
&GPT 4o &\cellcolor[HTML]{7ccaa4}0.90 &\cellcolor[HTML]{57bb8a}1.00 &\cellcolor[HTML]{57bb8a}1.00 &\cellcolor[HTML]{57bb8a}1.00 &\cellcolor[HTML]{66c195}0.96 &\cellcolor[HTML]{57bb8a}1.00 &\cellcolor[HTML]{57bb8a}1.00 &\cellcolor[HTML]{57bb8a}1.00 &\cellcolor[HTML]{57bb8a}1.00 &\cellcolor[HTML]{5fbe90}0.98 &\cellcolor[HTML]{57bb8a}1.00 \\
&GPT 4o CoT &\cellcolor[HTML]{57bb8a}1.00 &\cellcolor[HTML]{57bb8a}1.00 &\cellcolor[HTML]{57bb8a}1.00 &\cellcolor[HTML]{57bb8a}1.00 &\cellcolor[HTML]{57bb8a}1.00 &\cellcolor[HTML]{57bb8a}1.00 &\cellcolor[HTML]{5fbe90}0.98 &\cellcolor[HTML]{57bb8a}1.00 &\cellcolor[HTML]{57bb8a}1.00 &\cellcolor[HTML]{57bb8a}1.00 &\cellcolor[HTML]{66c195}0.96 \\
&Stateless &\cellcolor[HTML]{57bb8a}1.00 &\cellcolor[HTML]{f7d7d4}0.48 &\cellcolor[HTML]{f1faf5}0.58 &\cellcolor[HTML]{efb0aa}0.42 &\cellcolor[HTML]{e6f5ee}0.61 &\cellcolor[HTML]{fdf8f8}0.53 &\cellcolor[HTML]{fcfefd}0.55 &\cellcolor[HTML]{ec9c95}0.39 &0.54 &\cellcolor[HTML]{eda39d}0.40 &\cellcolor[HTML]{fcf1f0}0.52 \\
&MFC &\cellcolor[HTML]{fefdfd}0.54 &\cellcolor[HTML]{fcf3f2}0.52 &\cellcolor[HTML]{fefaf9}0.53 &\cellcolor[HTML]{fae9e8}0.51 &\cellcolor[HTML]{fcf0ef}0.52 &\cellcolor[HTML]{fefaf9}0.53 &\cellcolor[HTML]{fdfefe}0.55 &\cellcolor[HTML]{fdf5f4}0.53 &\cellcolor[HTML]{fcf0ef}0.52 &0.54 &\cellcolor[HTML]{f8fdfa}0.56 \\

\midrule\midrule

\multirow{12}{*}{Hands} &GPT 3.5 &\cellcolor[HTML]{57bb8a}1.00 &\cellcolor[HTML]{c1e6d4}0.62 &\cellcolor[HTML]{f1faf5}0.45 &\cellcolor[HTML]{e7847b}0.25 &\cellcolor[HTML]{ea948d}0.27 &\cellcolor[HTML]{f1b6b1}0.31 &\cellcolor[HTML]{f7d7d4}0.35 &\cellcolor[HTML]{f4c6c3}0.33 &\cellcolor[HTML]{efada8}0.30 &\cellcolor[HTML]{f7d7d4}0.35 &\cellcolor[HTML]{fcf0ef}0.38 \\
&GPT 3.5 CoT &\cellcolor[HTML]{57bb8a}1.00 &\cellcolor[HTML]{84ceaa}0.84 &\cellcolor[HTML]{ebf7f1}0.47 &\cellcolor[HTML]{fdf8f8}0.39 &\cellcolor[HTML]{fae8e6}0.37 &\cellcolor[HTML]{f7d7d4}0.35 &\cellcolor[HTML]{f7d7d4}0.35 &\cellcolor[HTML]{ea948d}0.27 &\cellcolor[HTML]{f7d7d4}0.35 &\cellcolor[HTML]{f6fcf9}0.43 &\cellcolor[HTML]{f9dfdd}0.36 \\
&Mixtral &\cellcolor[HTML]{57bb8a}1.00 &\cellcolor[HTML]{b6e2cd}0.66 &\cellcolor[HTML]{e6f5ee}0.49 &\cellcolor[HTML]{fdf8f8}0.39 &\cellcolor[HTML]{f7d7d4}0.35 &\cellcolor[HTML]{fae8e6}0.37 &\cellcolor[HTML]{f2beba}0.32 &\cellcolor[HTML]{efada8}0.30 &\cellcolor[HTML]{efada8}0.30 &\cellcolor[HTML]{e67c73}0.24 &\cellcolor[HTML]{f9fdfb}0.42 \\
&Mixtral CoT &\cellcolor[HTML]{57bb8a}1.00 &\cellcolor[HTML]{95d4b5}0.78 &\cellcolor[HTML]{dbf1e6}0.53 &\cellcolor[HTML]{fcf0ef}0.38 &\cellcolor[HTML]{fcf0ef}0.38 &\cellcolor[HTML]{f6fcf9}0.43 &\cellcolor[HTML]{f5cfcb}0.34 &\cellcolor[HTML]{f5cfcb}0.34 &\cellcolor[HTML]{ea948d}0.27 &\cellcolor[HTML]{f2beba}0.32 &\cellcolor[HTML]{fcf0ef}0.38 \\
&Llama3 8B &\cellcolor[HTML]{57bb8a}1.00 &\cellcolor[HTML]{e3f4ec}0.50 &\cellcolor[HTML]{f4fbf7}0.44 &\cellcolor[HTML]{f2beba}0.32 &\cellcolor[HTML]{e3f4ec}0.50 &\cellcolor[HTML]{ec9d96}0.28 &\cellcolor[HTML]{f9fdfb}0.42 &\cellcolor[HTML]{efada8}0.30 &\cellcolor[HTML]{f2beba}0.32 &\cellcolor[HTML]{d8efe4}0.54 &\cellcolor[HTML]{f9dfdd}0.36 \\
&Llama3 8B - CoT &\cellcolor[HTML]{57bb8a}1.00 &\cellcolor[HTML]{b6e2cd}0.66 &\cellcolor[HTML]{f9dfdd}0.36 &\cellcolor[HTML]{f9fdfb}0.42 &\cellcolor[HTML]{f5cfcb}0.34 &\cellcolor[HTML]{f9dfdd}0.36 &\cellcolor[HTML]{efada8}0.30 &\cellcolor[HTML]{f5cfcb}0.34 &\cellcolor[HTML]{f5cfcb}0.34 &\cellcolor[HTML]{f9dfdd}0.36 &\cellcolor[HTML]{f5cfcb}0.34 \\
&Llama3 70B &\cellcolor[HTML]{57bb8a}1.00 &\cellcolor[HTML]{57bb8a}1.00 &\cellcolor[HTML]{cdebdc}0.58 &\cellcolor[HTML]{fcf0ef}0.38 &\cellcolor[HTML]{fcf0ef}0.38 &\cellcolor[HTML]{f2beba}0.32 &\cellcolor[HTML]{f5cfcb}0.34 &\cellcolor[HTML]{f2beba}0.32 &\cellcolor[HTML]{e67c73}0.24 &\cellcolor[HTML]{f9dfdd}0.36 &\cellcolor[HTML]{e67c73}0.24 \\
&Llama3 70B - CoT &\cellcolor[HTML]{57bb8a}1.00 &\cellcolor[HTML]{57bb8a}1.00 &\cellcolor[HTML]{a6dbc1}0.72 &\cellcolor[HTML]{7fcba6}0.86 &\cellcolor[HTML]{73c79e}0.90 &\cellcolor[HTML]{8ad0ad}0.82 &\cellcolor[HTML]{84ceaa}0.84 &\cellcolor[HTML]{8ad0ad}0.82 &\cellcolor[HTML]{8ad0ad}0.82 &\cellcolor[HTML]{c7e9d8}0.60 &\cellcolor[HTML]{c1e6d4}0.62 \\
&GPT 4o &\cellcolor[HTML]{57bb8a}1.00 &\cellcolor[HTML]{57bb8a}1.00 &\cellcolor[HTML]{68c296}0.94 &\cellcolor[HTML]{57bb8a}1.00 &\cellcolor[HTML]{57bb8a}1.00 &\cellcolor[HTML]{57bb8a}1.00 &\cellcolor[HTML]{57bb8a}1.00 &\cellcolor[HTML]{57bb8a}1.00 &\cellcolor[HTML]{57bb8a}1.00 &\cellcolor[HTML]{73c79e}0.90 &\cellcolor[HTML]{57bb8a}1.00 \\
&GPT 4o CoT &\cellcolor[HTML]{57bb8a}1.00 &\cellcolor[HTML]{57bb8a}1.00 &\cellcolor[HTML]{5dbe8e}0.98 &\cellcolor[HTML]{57bb8a}1.00 &\cellcolor[HTML]{57bb8a}1.00 &\cellcolor[HTML]{57bb8a}1.00 &\cellcolor[HTML]{57bb8a}1.00 &\cellcolor[HTML]{57bb8a}1.00 &\cellcolor[HTML]{57bb8a}1.00 &\cellcolor[HTML]{57bb8a}1.00 &\cellcolor[HTML]{57bb8a}1.00 \\
&Stateless &\cellcolor[HTML]{57bb8a}1.00 &\cellcolor[HTML]{f7d7d4}0.35 &0.40 &\cellcolor[HTML]{fae8e6}0.37 &\cellcolor[HTML]{f4c6c3}0.33 &0.40 &\cellcolor[HTML]{f1b6b1}0.31 &\cellcolor[HTML]{f5cfcb}0.34 &\cellcolor[HTML]{efada8}0.30 &\cellcolor[HTML]{eda59f}0.29 &\cellcolor[HTML]{f5cfcb}0.34 \\
&MFC &\cellcolor[HTML]{f9e3e2}0.37 &\cellcolor[HTML]{fdfffe}0.41 &\cellcolor[HTML]{fcfefd}0.41 &\cellcolor[HTML]{fae8e6}0.37 &\cellcolor[HTML]{f9e3e2}0.37 &\cellcolor[HTML]{fae8e6}0.37 &\cellcolor[HTML]{fbeceb}0.38 &\cellcolor[HTML]{f7d7d4}0.35 &\cellcolor[HTML]{fcf0ef}0.38 &\cellcolor[HTML]{fefcfc}0.40 &\cellcolor[HTML]{f6fcf9}0.43 \\

\midrule\midrule

\multirow{12}{*}{Lights} &GPT 3.5 &\cellcolor[HTML]{70c59b}0.96 &\cellcolor[HTML]{fbefed}0.67 &\cellcolor[HTML]{efaba6}0.46 &\cellcolor[HTML]{f5cbc8}0.56 &\cellcolor[HTML]{f2bbb7}0.51 &\cellcolor[HTML]{ec9f98}0.42 &\cellcolor[HTML]{f3c2be}0.53 &\cellcolor[HTML]{ec9b95}0.41 &\cellcolor[HTML]{f1b8b3}0.50 &\cellcolor[HTML]{ea928a}0.38 &\cellcolor[HTML]{eea8a2}0.45 \\
&GPT 3.5 CoT &\cellcolor[HTML]{57bb8a}1.00 &\cellcolor[HTML]{e7f6ef}0.76 &\cellcolor[HTML]{f6d2cf}0.58 &\cellcolor[HTML]{fbebea}0.66 &\cellcolor[HTML]{fae5e3}0.64 &\cellcolor[HTML]{f9e2e0}0.63 &\cellcolor[HTML]{f8dbd9}0.61 &\cellcolor[HTML]{f6d2cf}0.58 &\cellcolor[HTML]{f4c8c4}0.55 &\cellcolor[HTML]{f9e2e0}0.63 &\cellcolor[HTML]{f3c2be}0.53 \\
&Mixtral &\cellcolor[HTML]{57bb8a}1.00 &0.72 &\cellcolor[HTML]{f9e2e0}0.63 &\cellcolor[HTML]{f4c8c4}0.55 &\cellcolor[HTML]{f1b8b3}0.50 &\cellcolor[HTML]{f0b5b0}0.49 &\cellcolor[HTML]{f0b2ad}0.48 &\cellcolor[HTML]{e67c73}0.31 &\cellcolor[HTML]{ea958e}0.39 &\cellcolor[HTML]{e98f87}0.37 &\cellcolor[HTML]{ec9b95}0.41 \\
&Mixtral CoT &\cellcolor[HTML]{57bb8a}1.00 &\cellcolor[HTML]{69c397}0.97 &\cellcolor[HTML]{a6dbc1}0.87 &\cellcolor[HTML]{bee5d2}0.83 &\cellcolor[HTML]{edf8f3}0.75 &\cellcolor[HTML]{e7f6ef}0.76 &\cellcolor[HTML]{f9fdfb}0.73 &0.72 &\cellcolor[HTML]{e1f3eb}0.77 &\cellcolor[HTML]{c9eada}0.81 &\cellcolor[HTML]{dbf1e6}0.78 \\
&Llama3 8B &\cellcolor[HTML]{57bb8a}1.00 &\cellcolor[HTML]{70c59b}0.96 &\cellcolor[HTML]{fdf8f8}0.70 &\cellcolor[HTML]{fbebea}0.66 &0.72 &\cellcolor[HTML]{fcf2f1}0.68 &\cellcolor[HTML]{f7d8d6}0.60 &\cellcolor[HTML]{f8dfdc}0.62 &\cellcolor[HTML]{f1b8b3}0.50 &\cellcolor[HTML]{f4fbf7}0.74 &\cellcolor[HTML]{fbebea}0.66 \\
&Llama3 8B - CoT &\cellcolor[HTML]{57bb8a}1.00 &\cellcolor[HTML]{93d4b4}0.90 &\cellcolor[HTML]{dbf1e6}0.78 &\cellcolor[HTML]{fae5e3}0.64 &\cellcolor[HTML]{fbebea}0.66 &\cellcolor[HTML]{f4fbf7}0.74 &\cellcolor[HTML]{f8dfdc}0.62 &\cellcolor[HTML]{f8dfdc}0.62 &\cellcolor[HTML]{f7d8d6}0.60 &\cellcolor[HTML]{fbebea}0.66 &\cellcolor[HTML]{f2bfba}0.52 \\
&Llama3 70B &\cellcolor[HTML]{57bb8a}1.00 &\cellcolor[HTML]{70c59b}0.96 &\cellcolor[HTML]{93d4b4}0.90 &\cellcolor[HTML]{87cfac}0.92 &\cellcolor[HTML]{abddc5}0.86 &\cellcolor[HTML]{9fd9bd}0.88 &\cellcolor[HTML]{9fd9bd}0.88 &\cellcolor[HTML]{f4fbf7}0.74 &\cellcolor[HTML]{c4e7d6}0.82 &\cellcolor[HTML]{f6d2cf}0.58 &\cellcolor[HTML]{fcf2f1}0.68 \\
&Llama3 70B - CoT &\cellcolor[HTML]{57bb8a}1.00 &\cellcolor[HTML]{57bb8a}1.00 &\cellcolor[HTML]{64c093}0.98 &\cellcolor[HTML]{64c093}0.98 &\cellcolor[HTML]{57bb8a}1.00 &\cellcolor[HTML]{64c093}0.98 &\cellcolor[HTML]{70c59b}0.96 &\cellcolor[HTML]{70c59b}0.96 &\cellcolor[HTML]{64c093}0.98 &\cellcolor[HTML]{57bb8a}1.00 &\cellcolor[HTML]{64c093}0.98 \\
&GPT 4o &\cellcolor[HTML]{57bb8a}1.00 &\cellcolor[HTML]{70c59b}0.96 &\cellcolor[HTML]{64c093}0.98 &\cellcolor[HTML]{57bb8a}1.00 &\cellcolor[HTML]{57bb8a}1.00 &\cellcolor[HTML]{57bb8a}1.00 &\cellcolor[HTML]{64c093}0.98 &\cellcolor[HTML]{57bb8a}1.00 &\cellcolor[HTML]{57bb8a}1.00 &\cellcolor[HTML]{57bb8a}1.00 &\cellcolor[HTML]{57bb8a}1.00 \\
&GPT 4o CoT &\cellcolor[HTML]{57bb8a}1.00 &\cellcolor[HTML]{57bb8a}1.00 &\cellcolor[HTML]{57bb8a}1.00 &\cellcolor[HTML]{57bb8a}1.00 &\cellcolor[HTML]{57bb8a}1.00 &\cellcolor[HTML]{57bb8a}1.00 &\cellcolor[HTML]{57bb8a}1.00 &\cellcolor[HTML]{57bb8a}1.00 &\cellcolor[HTML]{57bb8a}1.00 &\cellcolor[HTML]{57bb8a}1.00 &\cellcolor[HTML]{64c093}0.98 \\
&Stateless &\cellcolor[HTML]{57bb8a}1.00 &\cellcolor[HTML]{f2bfba}0.52 &\cellcolor[HTML]{f5cbc8}0.56 &\cellcolor[HTML]{f3c2be}0.53 &\cellcolor[HTML]{f1b8b3}0.50 &\cellcolor[HTML]{efafa9}0.47 &\cellcolor[HTML]{f0b5b0}0.49 &\cellcolor[HTML]{ea958e}0.39 &\cellcolor[HTML]{f3c2be}0.53 &\cellcolor[HTML]{f8dbd9}0.61 &\cellcolor[HTML]{eda59f}0.44 \\
&MFC &\cellcolor[HTML]{f6d2cf}0.58 &\cellcolor[HTML]{fae8e7}0.65 &\cellcolor[HTML]{f5cbc8}0.56 &\cellcolor[HTML]{f2bdb9}0.52 &\cellcolor[HTML]{f2bbb7}0.51 &\cellcolor[HTML]{f4c7c3}0.54 &\cellcolor[HTML]{f4c5c1}0.54 &\cellcolor[HTML]{f5cbc8}0.56 &\cellcolor[HTML]{f3c2be}0.53 &\cellcolor[HTML]{f7dad7}0.61 &\cellcolor[HTML]{f4c5c1}0.54 \\
\bottomrule
\end{tabular}}
    \caption{The main evaluation results of systems on different tasks reported in terms of accuracy at various depths. \textbf{Linear} and \textbf{Hands} refer to \texttt{LinearWorld} and \texttt{HandSwap} tasks respectively. Also, systems with Chain-of-Thought integrated in their evaluation are indicated with "CoT" in their names.}
    \label{tab:main_results}
\end{table*}

\noindent \textbf{Impact of Depth.} The first observation is the effect of state tracking depth on system performance. The results indicate two distinct behaviors: smaller models (such as the LLama3 8B variant) and older models (GPT-3.5 and Mixtral) experience a significant drop in performance once the depth reaches 2, in contrast to larger and newer models (LLama3 70B variant and GPT-4o) which maintain acceptable accuracy even at greater depths, particularly when combined with CoT.

Interestingly, almost all LLMs (with the exception of LLama3 8B) have near-perfect accuracy at depth 0 over all the tasks, where no update statements are present. This suggests that the models can comprehend the natural language version of the task provided in prompts. They only struggle when it comes to following the update statements, due to the lack of access to on-demand memory.

\noindent \textbf{Impact of Chain of Thought.} Another evident observation is the benefit of incorporating Chain of Thought (CoT) into the model prompting process. With CoT, systems can use the input context as a form of memory to store intermediate states, thereby making it easier to solve the tasks at hand and follow the intermediate states of the evolving underlying environment.

Figure \ref{fig:cot_impact} shows the difference in the performance  averaged across all models, excluding the two best-performing ones (LLama3 70B and GPT-4), when integrated with CoT compared to when CoT is not employed. The trend clearly highlights the benefit of incorporating CoT into the process.

\noindent \textbf{Stateful Vs Stateless.} One possible question is the extent to which LLMs rely on the initial state provided to them, potentially disregarding any update statements and solely using the initial state to answer the query. To this end, we designed a control stateless baseline, that is simply a system that disregards all update statements and answers each query solely based on the provided initial state.

The results illustrated in Table \ref{tab:main_results} suggest that although not perfect, LLMs are in general able to answer queries about the updated environment more accurately than a naive stateless system. This observation indicates that the models which fail at greater depths, still follow the state to some extent.


\begin{figure*}[!htbp]

        \includegraphics[width=\linewidth]{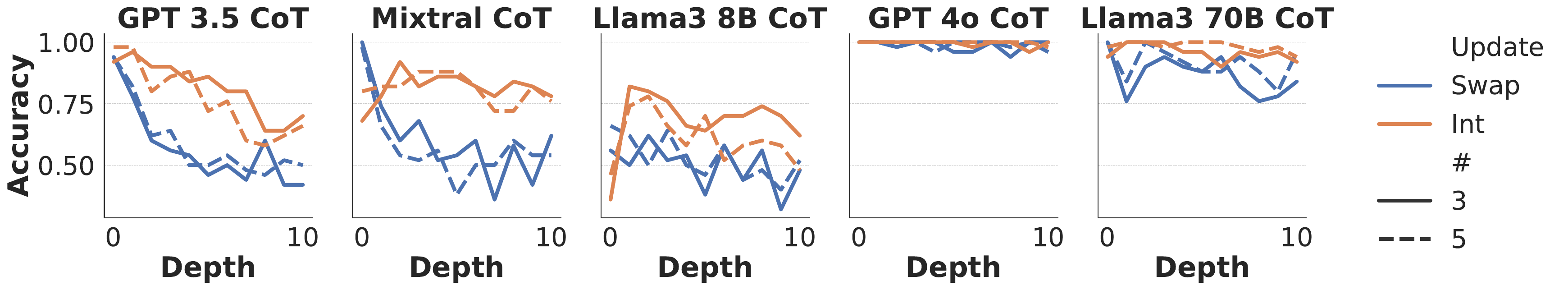}
        
    \caption{Accuracy at different depths comparing the "swap" update type with the "integer" update type in the \texttt{LinearWorld} task (the state-dependent query variants) for all CoT integrated systems.}
    \label{fig:utype-linear}

\end{figure*}


\section{Analysis}
\label{sec:complementary_study}

We present additional control experiments and analyses to complement the main results presented above. 
 The experiments in this section are conducted on the \texttt{LinearWorld} task due to its simplicity and flexibility to study alternative configurations.

\subsection{Control Experiments}

The control experiments and the main findings are summarized below. In the appendix (Tables \ref{tab:complementary_open} and \ref{tab:complementary_gpt}), we include the full experimental results.

\noindent \textbf{Number of Individuals.} Increasing the number of entities (in this case people) in the configuration of \texttt{LinearWorld} task, adds to the complexity of the choice space for the model; the larger the number of entities gets, the more possible configurations exist in which they can stand in a line, thus increasing the number of possible states the model must track. With that in mind, we designed a version of the \texttt{LinearWorld} task with 5 entities standing in a line, instead of the original 3 entities. 

Contrary to expectations, the models' performance does not drop when more entities are introduced to the problem. One possible explanation is that with more individuals, there are more possibilities to choose two random entities for the final query whose relative positions remain unaffected by the update statement. In that case, models could simply refer to the initial state of the entities to answer the query. We will see further evidence of this behavior when comparing the "state-dependent" and "random" query types, as discussed next.


\noindent \textbf{State Dependant Queries.} One consequence of having a greater number of individuals is that the chosen query more often asks about the relative position of two entities whose positions have not been modified by the updates. This, in turn makes the task easier for the model by simply answering the queries solely based on the initial state. In order to prevent that, we introduce a setting in which the person placement of each query is chosen so that the relative positions of entities in question differ from their initial state. The results for this setting, including a figure plotting the results by depth (figure \ref{fig:qtype-linear.pdf}), are shown in the appendix due to space constraints. In general, the results follow a similar trend in this and the default setting. However, the performance gap between query types is most pronounced in models that excel at lower depths but struggle at higher depths, with the default query type (random) being the easier of the two.

\noindent \textbf{Integer Displacement.} In order to assess the impact of various update types, we introduce a variant of \texttt{LinearWorld} task, in which the update statements are simply the displacement of one of individuals by an arbitrary integer value (e.g. "Entity A moves 5 units to the left"). Following the update statements presented in this manner, requires a minimal mathematical calculation to determine the final positions of entities as a first step, and then deducing their relative positions to answer the query. One possibility is that models perform mathematical operations by simply memorizing the results of summing two small integers that have appeared in their pre-training data. To make this less plausible, we introduce initial state positions and displacements with large integer values as well, reducing the likelihood that these specific number combinations have occurred in the pre-training data.


As illustrated in Figure \ref{fig:utype-linear}, systems surprisingly exhibit superior performance in the integer setting compared to swap. A subsequent qualitative analysis revealed that struggling systems fail to make logical statements and update the state coherently after receiving swap updates. In contrast, models tend to follow state changes more logically in the integer update setting. Some examples of errors can be found in the Appendix. A key difference between swap and integer displacement update types is revealed upon inspecting the responses: models attempt to isolate updates regarding each entity from all other updates, then aggregate and solve them out of sequence (e.g. $5+3-2=6$ as to solve the position of one person after three updates instead of performing each update separately).

\subsection{Proficiency in Mathematical Operations}

To assess how models perform when introduced to mathematical operations, we analyze their responses through a simple approach. We search for simple mathematical expressions in the model outputs using basic regex matching, and then we compute the percentage of correctly calculated mathematical expressions by the model.

\begin{figure}[!htbp]

        \includegraphics[width=\linewidth]{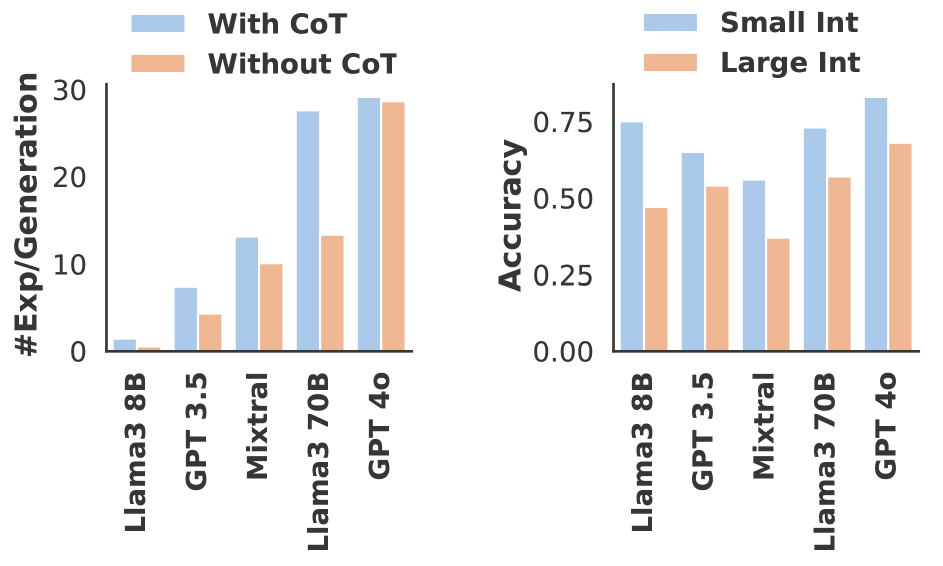}

    
    \caption{Average number of mathematical expressions per model response for all the models (left), and accuracy of generated expression evaluations across models integrated with CoT (right).}
    \label{fig:math_eval}
\end{figure}

Figure \ref{fig:math_eval} presents the accuracy of the model's calculations as well as the average number of mathematical expressions written in the output response by the model per generation. GPT-4 demonstrates superior performance not only in mathematical calculations but also in generating a higher number of expressions per output. Moreover, the difference between the number of expressions in CoT versus non-CoT settings is less pronounced for GPT-4 (we will analyse this specific behaviour in the following section).
In contrast, other models produce fewer expressions when CoT is not employed and generally exhibit lower accuracy in their calculations. This difference is particularly reflected in the results when comparing the performance of the large variant of Llama3 on integer-based tasks with and without CoT prompting.

Notably, the models exhibit decreased calculation accuracy when presented with large integer numbers compared to smaller ones. However, this deficiency is less marked in the performance of the models on the \texttt{LinearWorld} task. This may be attributed to the fact that in scenarios involving large integers, there is more room for calculation errors while still maintaining the correct relative order of individuals.

\subsection{Response Length}

As a basic proxy to measure how often models employ CoT or other sort of reasoning in their responses (whether instructed to do so or otherwise), we calculate Pearson's correlation coefficient between the model response length and the depth of the query. A larger correlation indicates that the system is engaging more in deeper queries, suggesting a higher likelihood of a reasoning process being articulated by the model.

The results reported in Table \ref{tab:cot_activation} suggest that better-performing models demonstrate a stronger correlation between response length and the depth at which the response is generated. This indicates a higher likelihood of Chain of Thought reasoning being activated in these systems. As expected, the CoT mechanism is more likely to be activated in systems that are explicitly instructed to employ it.

Interestingly, a comparison of CoT activation across different systems reveals that for GPT-4, the CoT mechanism is activated to a similar degree regardless of explicit instructions. This suggests that GPT-4 employs CoT reasoning intrinsically, even when not specifically prompted to articulate its reasoning process in the output. 
These findings align with the observation that GPT is more likely to write mathematical expressions in the tasks with integer-based update type, even without explicit instructions regarding the activation of CoT, as discussed in the previous section.

\begin{table}[h!]
    \tiny
    \resizebox{\linewidth}{!}{\setlength{\tabcolsep}{12pt}
\begin{tabular}{lrrrrr}\toprule
\textbf{System} &\textbf{Pearson $r$} \\\midrule 
Llama3 8B / + CoT &0.01 / 0.12 \\
GPT 3.5 / + CoT &0.03 / 0.04\\
Mixtral / + CoT&0.32 / 0.46\\
Llama3 70B / + CoT &0.35 / 0.79\\
GPT 4o / + CoT &0.85 / 0.91\\
\bottomrule
\end{tabular}}
    \caption{Pearson correlation between response length and generation depth across models.}
    \label{tab:cot_activation}
\end{table}

\section{Conclusion}
In this paper, we investigated the problem of state tracking in LLMs, specifically their ability to maintain state through sequential and repetitive updates by conducting a set of controlled experiments. For this, we have constructed a new benchmark based on three related tasks that are specifically designed to test the state tracking capabilities of LLMs. Our experiments indicate that the newer generation of models, such as GPT-4 and Llama3 70B (the larger variant of the Llama3 models), can effectively maintain state by keeping intermediate states, especially when integrated with the chain of thought reasoning technique. In contrast, the smaller variant of the Llama3 model, Llama3 8B, and older generation LLMs, such as Mixtral and GPT-3.5, struggle to utilize their input context as a form of inference-time note-keeping memory for state tracking after a certain number of update statements.


\section*{Limitations}


Despite the comprehensive nature of our study, several limitations must be acknowledged. First, our experiments were conducted on a limited set of models which may not fully capture the broader capabilities and nuances of state tracking across different architectures and sizes. Second, the prompts used in our experiments were also limited, which could affect the generalizability of our findings. Variations in prompt design might yield different results, highlighting the need for further exploration with diverse prompting strategies. Additionally, some of our analysis was confined to a single task, which may not provide a comprehensive understanding of state tracking across different contexts. Furthermore, we limited our experiments to a maximum depth of 10 update statements. This restriction might overlook the models' performance in handling deeper sequences of updates.
\end{spacing}

\bibliography{custom}

\appendix

\label{sec:appendix}

\section{Prompt Examples}
\label{app:prompt_examples}

In this section, we present examples of prompts used in our experiments. The examples provided here employ CoT. As for the simple non-CoT versions of these prompts, we simply omit the \texttt{Reasoning} parts, and the instructions are adjusted accordingly to not request the model's reasoning process.

\textbf{Note:} The red italicized comments are not part of the actual input given to the model but serve to clarify various aspects of the prompts for the reader.

\subsection{LinearWorld}
\label{app:prompt_examples:linear}

\begin{promptbox}
\promptcomment{--- This is the instruction we provide for the model: ---}

\vspace{2mm}
\noindent Consider a 2D world with three individuals, each starting at a given initial position, and a series of movements for each person, where `left' signifies a move towards negative coordinates and `right' denotes a move towards positive coordinates. Your task is to calculate and report the final position of each person after the specified sequence of movements and respond to a given query about their respective positions. Your response should strictly follow this format: `Answer: [[response]]', the response in the double-closed brackets is either left or right, e.g, `Answer: [[left]]' and `Answer: [[right]]'. Here is an example:

\vspace{2mm}
\promptcomment{--- Here we provide an example for the model: ---}

\vspace{2mm}
\noindent Person A starts at position 0\\
Person B starts at position 1\\
Person C starts at position 2

\begin{itemize}
    \item Person C swaps position with Person A.
    \item Person A swaps position with Person B.
\end{itemize}

\noindent Query: Where does Person B stand with respect to Person C now?

\noindent Options: left , right

\noindent Reasoning: Considering all the previous movements, Person B is at position 2. Person C is at position 0. Since 2 is greater than 0, Person B is to the right of Person C.

\noindent Answer: [[right]]

\vspace{2mm}
\promptcomment{--- Now we provide the actual test instance: ---}

\vspace{2mm}
\noindent Now your turn:

\noindent Person A starts at position 0\\
Person B starts at position 1\\
Person C starts at position 2

\noindent Person B swaps position with Person C.\\
Person C swaps position with Person A.\\
\promptcomment{(Note that this instance has a depth of 2)}

\noindent Query: Where does Person B stand with respect to Person A now?

\noindent Options: left , right

\noindent Reasoning:
\end{promptbox}

\subsection{HandSwap}
\label{app:prompt_examples:hands}

\begin{promptbox}
\promptcomment{--- This is the instruction we provide for the model: ---}

\vspace{2mm}
\noindent Consider three individuals, each holding an object. Given initial object that each individual holds, and a series of swaps with other individuals, your task is to respond to a given query about the final object that one of the individuals is holding. Your response should strictly follow this format: `Answer: [[response]]', with the response in the double-closed brackets chosen from this list of the options: [apple , banana , nothing], e.g, `Answer: [[apple]]' or `Answer: [[banana]]' or `Answer: [[nothing]]'. Here is an example:

\vspace{2mm}
\promptcomment{--- Here we provide an example for the model: ---}

\vspace{2mm}
\noindent Person A is initially holding apple\\
Person B is initially holding banana\\
Person C is initially holding nothing

\noindent Person B swaps hands with Person C.\\
Person C swaps hands with Person B.

\noindent Query: What is in the hands of Person B now?

\noindent Options: apple , banana , nothing

\noindent Reasoning: Considering all the previous swaps, Person B is now holding banana.

\noindent Answer: [[banana]]

\vspace{2mm}
\promptcomment{--- Now we provide the actual test instance: ---}

\vspace{2mm}
\noindent Now your turn:

\noindent Person A is initially holding apple\\
Person B is initially holding banana\\
Person C is initially holding nothing

\noindent Person C swaps hands with Person B.\\
Person A swaps hands with Person C.\\
Person C swaps hands with Person A.\\
\promptcomment{(Note that this instance has a depth of 3)}

\noindent Query: What is in the hands of Person C now?

\noindent Options: apple , banana , nothing

\noindent Reasoning:
\end{promptbox}

\subsection{Lights}
\label{app:prompt_examples:lights}

\begin{promptbox}
\promptcomment{--- This is the instruction we provide for the model: ---}

\vspace{2mm}
\noindent Consider a scenario where there are a set of switches and multiple rooms each equipped with a light. Each switch affects the state of certain lights when pressed. Given the initial state of the lights in each room and the effects of pressing each switch, your task is to respond to queries about the final state of a specific room after a sequence of switch presses. Your response format must strictly follow the following pattern: `Answer: [[response]]', where the response enclosed in double brackets indicates whether the light is on or off, for instance, `Answer: [[on]]' or `Answer: [[off]]'. Here is an example:

\vspace{2mm}
\promptcomment{--- Here we provide an example for the model: ---}

\vspace{2mm}
\noindent The light in room A is initially off.\\
The light in room B is initially off.\\
The light in room C is initially off.

\noindent When switch 1 is pressed, it turns the lights of room A on.\\
When switch 1 is pressed, it turns the lights of room B off.\\
When switch 2 is pressed, it turns the lights of room B on.\\
When switch 2 is pressed, it turns the lights of room C off.\\
When switch 3 is pressed, it turns the lights of room C on.\\
When switch 3 is pressed, it turns the lights of room A off.\\
When switch 4 is pressed, it turns the lights of room A, room B and room C on.\\
When switch 5 is pressed, it turns the lights of room A, room B and room C off.

\noindent The switch 3 is pressed.\\
The switch 3 is pressed.

\noindent Query: What is the final light status in room A?

\noindent Options: on, off

\noindent Reasoning: Starting from the first key press: The switch 3 is pressed and it turns the light in room A off. The switch 3 is pressed and it turns the light in room A off. Therefore the final status of the light in room A is off.

\noindent Answer: [[off]]

\vspace{2mm}
\promptcomment{--- Here we provide the actual test instance: ---}

\vspace{2mm}
\noindent Now your turn:

\noindent The light in room A is initially off.\\
The light in room B is initially off.\\
The light in room C is initially off.

\noindent When switch 1 is pressed, it turns the lights of room A on.\\
When switch 1 is pressed, it turns the lights of room B off.\\
When switch 2 is pressed, it turns the lights of room B on.\\
When switch 2 is pressed, it turns the lights of room C off.\\
When switch 3 is pressed, it turns the lights of room C on.\\
When switch 3 is pressed, it turns the lights of room A off.\\
When switch 4 is pressed, it turns the lights of room A, room B and room C on.\\
When switch 5 is pressed, it turns the lights of room A, room B and room C off.

\noindent The switch 3 is pressed.\\
The switch 5 is pressed.\\
The switch 5 is pressed.\\
The switch 1 is pressed.\\
\promptcomment{(Note that this instance has a depth of 4)}

\noindent Query: What is the final light status in room C?

\noindent Options: on, off

\noindent Reasoning:
\end{promptbox}

\section{Model Response Examples}
\label{app:response_examples}

In this section, we present examples of model responses in instances where errors occur. Note that a simplified version of the input to the model has been provided for each response. However, the models do not receive the input in this simplified format. For the full template in which these inputs are presented, please refer to section \ref{app:prompt_examples}.

\textbf{Note:} The red italicized comments are not part of the actual model responses but serve to clarify various errors of the systems for the reader.

\subsection{Llama3 8B}
\label{app:response_examples:llama_small}

\paragraph{Integer.} Sample input:

\begin{inputbox}
A starts at 1.\\
B starts at 2.\\
C starts at 3.

B moves 2 to right.\\
C moves 5 to left.\\
A moves 5 to right.\\
B moves 5 to right.\\
A moves 9 to left.\\
C moves 3 to left.\\
A moves 15 to right.\\
B moves 8 to right.\\
A moves 20 to left.

Where does B stand with respect to C?
\end{inputbox}

\noindent Sample response:

\begin{responsebox}
Person A starts at position 1 Person B starts at position 2 Person C starts at position 3

Person B moves 2 units to right.\\
-> Person B is now at position 4

Person C moves 5 units to left.\\
-> Person C is now at position -2

Person A moves 5 units to right.\\
-> Person A is now at position 6

Person B moves 5 units to right.\\
-> Person B is now at position 9

Person A moves 9 units to left.\\
-> Person A is now at position -3

Person C moves 3 units to left.\\
-> Person C is now at position 6 \errorcomment{Wrong!}

Person A moves 15 units to right.\\
-> Person A is now at position 12

Person B moves 8 units to right.\\
-> Person B is now at position 5 \errorcomment{Wrong!}

Person A moves 20 units to left.\\
-> Person A is now at position -8

Query: Where does Person B stand with respect to Person C now? Options: left, right Answer: right
\end{responsebox}

\paragraph{Swap.} Sample input:

\begin{inputbox}
A starts at 0.\\
B starts at 1.\\
C starts at 2.

C swaps with A.\\
A swaps with B.\\
B swaps with A.\\
B swaps with C.\\
B swaps with A.\\
C swaps with A.\\
C swaps with B.\\
C swaps with B.

Where does B stand with respect to C?
\end{inputbox}

\noindent Sample response:

\begin{responsebox}
After considering all the previous movements, Person B's position is as follows:

Person B is at position 0 :\\
 - initially at 1,\\
 - then swapped with Person A,\\
 - then swapped with Person C,\\
 - then swapped with Person A again,\\
 - and finally swapped with Person B\\
 \errorcomment{(Wrong: B can not swap with B!)}

Person C is at position 1 :\\
 - initially at 2,\\
 - then swapped with Person A,\\
 - then swapped with Person B,\\
 - then swapped with Person A,\\
 - then swapped with Person B,\\
 - and finally swapped with Person C \errorcomment{(Wrong again: C can not swap with C)}

Since 0 is less than 1, Person B is to the left of Person C. Answer: [[left]]"
\end{responsebox}

\subsection{Llama3 70B}
\label{app:response_examples:llama_large}

\paragraph{Integer.} Sample input:

\begin{inputbox}
A starts at 1.\\
B starts at 2.\\
C starts at 3.

B moves 2 to left.\\
A moves 7 to right.\\
C moves 8 to right.\\
A moves 4 to right.\\
C moves 3 to right.\\
A moves 5 to right.\\
C moves 16 to left.\\
B moves 5 to left.\\
C moves 20 to right.\\
A moves 2 to right.

Where does B stand with respect to A?
\end{inputbox}

\noindent Sample response:

\begin{responsebox}
Person A:\\
- starts at position\\
- moves 7 units to right: 1 + 7 = 8\\
- moves 4 units to right: 8 + 4 = 12\\
- moves 5 units to right: 12 + 5 = 17\\
- moves 2 units to right: 17 + 2 = 19

Person B:\\
- starts at position 2\\
- moves 2 units to left: 2 - 2 = 0\\
- moves 5 units to left: 0 - 5 = -5

Person C:\\
- starts at position 3\\
- moves 8 units to right: 3 + 8 = 11\\
- moves 3 units to right: 11 + 3 = 14\\
- moves 16 units to left: 14 - 16 = -2\\
- moves 20 units to right: -2 + 20 = 18

Now, let's compare the final positions of Person A and Person B: Person A is at position 19 Person B is at position -5 Since 19 is greater than -5, Person A is to the right of Person B.

Answer: [[right]]\\
\errorcomment{Wrong since the query asks for position of person}\\
\errorcomment{B with respect to A not the other way around}
\end{responsebox}

\paragraph{Swap.} Sample input:

\begin{inputbox}
A starts at 0.\\
B starts at 1.\\
C starts at 2.

A swaps with C.\\
A swaps with B.\\
B swaps with A.\\
C swaps with B.\\
C swaps with A.\\
A swaps with C.\\
A swaps with C.\\
A swaps with C.\\
B swaps with A.\\
A swaps with B.

Where does C stand with respect to A?
\end{inputbox}

\noindent Sample response:

\begin{responsebox}
1. Person A starts at position 0, Person B starts at position 1, and Person C starts at position 2.\\
2. Person A swaps position with Person C, so Person A is now at position 2, and Person C is at position 0.\\
3. Person A swaps position with Person B, so Person A is now at position 1, and Person B is at position 2.\\
4. Person B swaps position with Person A, so Person B is now at position 1, and Person A is at position 2.\\
5. Person C swaps position with Person B, so Person C is now at position 1, and Person B is at position 0.\\
6. Person C swaps position with Person A, so Person C is now at position 2, and Person A is at position 1.\\
\errorcomment{(the following is wrong!)}\\
7. Person A swaps position with Person C (three times), but this doesn't change their positions, so Person A remains at position 1, and Person C remains at position 2.\\
8. Person B swaps position with Person A, so Person B is now at position 1, and Person A is at position 0.\\
9. Person A swaps position with Person B, so Person A is now at position 1, and Person B is at position 0.

After all the movements, Person A is at position 1, and Person C is at position 2. Since 1 is smaller than 2, Person C is to the right of Person A. Answer: [[right]]"
\end{responsebox}

\subsection{Mixtral}
\label{app:response_examples:Mixtral}

\paragraph{Integer.} Sample input:

\begin{inputbox}
A starts at 1.\\
B starts at 2.\\
C starts at 3.

B moves 6 to right.\\
C moves 4 to left.\\
A moves 6 to left.\\
C moves 6 to left.\\
A moves 16 to right.\\
B moves 4 to right.\\
A moves 20 to left.\\
C moves 4 to left.\\
A moves 26 to right.\\
B moves 25 to left.

Where does C stand with respect to A?
\end{inputbox}

\noindent Sample response:

\begin{responsebox}
To answer this question, we first need to determine the current positions of Persons A and C after the given sequence of movements.

Person A's movements:\\
* Starts at position 1\\
* Moves 6 units to the left, making the new position 1 - 6 = -5\\
 \errorcomment{(the following is wrong!)}\\
* Moves 6 units to the left, making the new position -5 - 6 = -11\\
* Moves 16 units to the right, making the new position -11 + 16 = 5\\
* Moves 20 units to the left, making the new position 5 - 20 = -15\\
* Moves 26 units to the right, making the new position -15 + 26 = 11

Person C's movements:\\
* Starts at position 3\\
* Moves 4 units to the left, making the new position 3 - 4 = -1\\
* Moves 6 units to the left, making the new position -1 - 6 = -7\\
* Moves 4 units to the left, making the new position -7 - 4 = -11\\
* Moves 6 units to the left, making the new position -11 - 6 = -17

Now that we have the final positions of Persons A and C, we can compare them:\\
* Person A is at position 11\\
* Person C is at position -17

Since 11 is greater than -17, Person A is to the right of Person C.\\
\errorcomment{The error in math did not matter,}\\
\errorcomment{model is answering the wrong query regardless.}\\
Therefore, the answer is: Answer: [[right]]"
\end{responsebox}

\paragraph{Swap.} Sample input:

\begin{inputbox}
A starts at 0.\\
B starts at 1.\\
C starts at 2.

A swaps with C.\\
B swaps with C.\\
B swaps with C.\\
B swaps with C.\\
A swaps with C.\\
C swaps with B.\\
B swaps with A.\\
B swaps with A.\\
C swaps with B.\\
A swaps with C.

Where does A stand with respect to C?
\end{inputbox}

\noindent Sample response:

\begin{responsebox}
\errorcomment{The model is making so many incorrect statements}\\
\errorcomment{since it is trying to aggregate all}\\
\errorcomment{swap statements for each entity.}

Person A starts at position 0,\\
after all the specified movements, A swaps positions with C twice, and  \errorcomment{(wrong!)}\\
B swaps positions with A three times. \errorcomment{(wrong!)}\\
This means A and B essentially swap positions. \errorcomment{(wrong!)}\\
So, since B ends up at position 0 (by swapping with A), \errorcomment{(correct}\\
\errorcomment{but by chance!)}\\
and C is at position 2,  \errorcomment{(wrong!)}\\
then A must be at position 1.\\
\errorcomment{>> logical conclusion based on wrong premises}\\
Therefore, with respect to Person C, who is at position 2, Person A is to the left. Answer: [[left]]"
\end{responsebox}






\begin{figure*}[!b]

        \includegraphics[width=\linewidth]{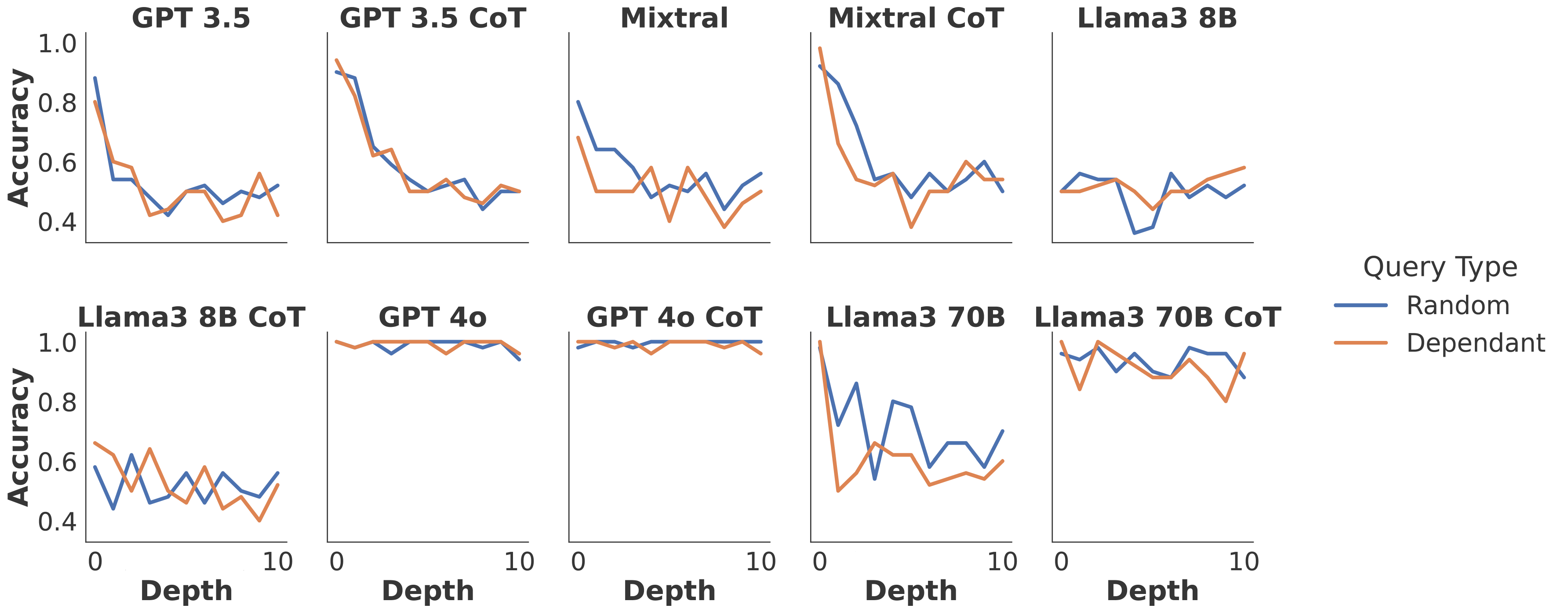}

    \caption{Accuracy at different depths comparing the "state-dependant" query type with the "random" query type in the \texttt{LinearWorld} task (the variants with 5 individuals) accross all the systems.}
    \label{fig:qtype-linear.pdf}

\end{figure*}

\section{Impact of Query Type.}

According to the results illustrated in Tables \ref{tab:complementary_open} and \ref{tab:complementary_gpt}, the query type has a greater impact on tasks involving five individuals compared to those with three. This may be attributed to the fact that a single update statement is more likely to alter the relative positions of all entities, thus inherently achieving state dependence without the need to enforce it.

Figure \ref{fig:qtype-linear.pdf} compares system performance on "state-dependent" versus "random" queries for tasks involving five entities. GPT-4o demonstrates near-perfect performance across both query types, emerging as the top-performing model. In contrast, the the least effective model being the small variant of LLaMA3 barely outperforms a random baseline. For these two extremes, the difference in performance between query types is negligible. However, the performance gap between query types is most pronounced in models that excel at lower depths but struggle at higher depths, particularly the systems based on Mixtral model and the large variant of LLaMA3. These models find the state-dependent setting more challenging, as expected.

\section{Detailed Result Tables}
In this section, a complete table of the results of our controlled experiments is presented.

\begin{table*}[!htbp]
    \resizebox{\linewidth}{!}{\begin{tabular}{llll cccccccccccc}\toprule
\multirow{2}{*}{System} &\multirow{2}{*}{\# } &\multirow{2}{*}{Query} &\multirow{2}{*}{Update} &\multicolumn{11}{c}{Depth} \\ \cmidrule(rl){5-15} 

& & & &0 &1 &2 &3 &4 &5 &6 &7 &8 &9 &10 \\ 

\midrule

\multirow{7}{*}{\begin{sideways} GPT 3.5\end{sideways}} &3 &Ran &Swap &\cellcolor[HTML]{6ec59a}0.96 &\cellcolor[HTML]{f6d1ce}0.54 &\cellcolor[HTML]{f3c0bc}0.48 &\cellcolor[HTML]{eea9a3}0.40 &\cellcolor[HTML]{efafa9}0.42 &\cellcolor[HTML]{f0b4af}0.44 &\cellcolor[HTML]{f4c6c2}0.50 &\cellcolor[HTML]{eea9a3}0.40 &\cellcolor[HTML]{f6d1ce}0.54 &\cellcolor[HTML]{f6d1ce}0.54 &\cellcolor[HTML]{f5cbc8}0.52 \\
&3 &Dep &Swap &\cellcolor[HTML]{e9f6f0}0.74 &\cellcolor[HTML]{f7d7d4}0.56 &\cellcolor[HTML]{f4c6c2}0.50 &\cellcolor[HTML]{f3c0bc}0.48 &\cellcolor[HTML]{eea9a3}0.40 &\cellcolor[HTML]{efafa9}0.42 &\cellcolor[HTML]{f7d7d4}0.56 &\cellcolor[HTML]{f8dcda}0.58 &\cellcolor[HTML]{f4c6c2}0.50 &\cellcolor[HTML]{f0b4af}0.44 &\cellcolor[HTML]{f3c0bc}0.48 \\
&3 &Dep &Int &\cellcolor[HTML]{a6dbc1}0.86 &\cellcolor[HTML]{c7e9d8}0.80 &\cellcolor[HTML]{a6dbc1}0.86 &\cellcolor[HTML]{8fd2b1}0.90 &\cellcolor[HTML]{d3ede0}0.78 &\cellcolor[HTML]{f4fbf8}0.72 &\cellcolor[HTML]{fcf3f2}0.66 &\cellcolor[HTML]{fdf9f8}0.68 &\cellcolor[HTML]{f4fbf8}0.72 &\cellcolor[HTML]{f8dcda}0.58 &\cellcolor[HTML]{fdf6f5}0.67 \\
&5 &Ran &Swap &\cellcolor[HTML]{9bd7b9}0.88 &\cellcolor[HTML]{f6d1ce}0.54 &\cellcolor[HTML]{f6d1ce}0.54 &\cellcolor[HTML]{f3c0bc}0.48 &\cellcolor[HTML]{efafa9}0.42 &\cellcolor[HTML]{f4c6c2}0.50 &\cellcolor[HTML]{f5cbc8}0.52 &\cellcolor[HTML]{f1bab5}0.46 &\cellcolor[HTML]{f4c6c2}0.50 &\cellcolor[HTML]{f3c0bc}0.48 &\cellcolor[HTML]{f5cbc8}0.52 \\
&5 &Dep &Swap &\cellcolor[HTML]{c7e9d8}0.80 &\cellcolor[HTML]{f9e2e0}0.60 &\cellcolor[HTML]{f8dcda}0.58 &\cellcolor[HTML]{efafa9}0.42 &\cellcolor[HTML]{f0b4af}0.44 &\cellcolor[HTML]{f4c6c2}0.50 &\cellcolor[HTML]{f4c6c2}0.50 &\cellcolor[HTML]{eea9a3}0.40 &\cellcolor[HTML]{efafa9}0.42 &\cellcolor[HTML]{f7d7d4}0.56 &\cellcolor[HTML]{efafa9}0.42 \\
&5 &Dep &Int &\cellcolor[HTML]{b1e0c9}0.84 &\cellcolor[HTML]{d3ede0}0.78 &\cellcolor[HTML]{fbedec}0.64 &0.70 &\cellcolor[HTML]{f9e2e0}0.60 &\cellcolor[HTML]{f7d7d4}0.56 &\cellcolor[HTML]{fae8e6}0.62 &\cellcolor[HTML]{fcf3f2}0.66 &\cellcolor[HTML]{f5cbc8}0.52 &\cellcolor[HTML]{fcf3f2}0.66 &\cellcolor[HTML]{f9e2e0}0.60 \\
&5 &Dep &L-Int &\cellcolor[HTML]{bce4d1}0.82 &\cellcolor[HTML]{fae8e6}0.62 &\cellcolor[HTML]{f4fbf8}0.72 &\cellcolor[HTML]{f8dcda}0.58 &\cellcolor[HTML]{fbedec}0.64 &\cellcolor[HTML]{f9e2e0}0.60 &\cellcolor[HTML]{fdf9f8}0.68 &\cellcolor[HTML]{fae8e6}0.62 &\cellcolor[HTML]{f7d7d4}0.56 &\cellcolor[HTML]{f6d1ce}0.54 &\cellcolor[HTML]{fbedec}0.64 \\

\midrule \midrule

\multirow{7}{*}{\begin{sideways} GPT 3.5 CoT \end{sideways}} &3 &Ran &Swap &\cellcolor[HTML]{63c092}0.98 &\cellcolor[HTML]{a6dbc1}0.86 &\cellcolor[HTML]{f7d7d4}0.56 &\cellcolor[HTML]{f9e2e0}0.60 &\cellcolor[HTML]{f3c0bc}0.48 &\cellcolor[HTML]{f0b4af}0.44 &\cellcolor[HTML]{f3c0bc}0.48 &\cellcolor[HTML]{eda39d}0.38 &\cellcolor[HTML]{f7d7d4}0.56 &\cellcolor[HTML]{f0b4af}0.44 &\cellcolor[HTML]{f3c0bc}0.48 \\
&3 &Dep &Swap &\cellcolor[HTML]{79c9a2}0.94 &\cellcolor[HTML]{d3ede0}0.78 &\cellcolor[HTML]{f9e2e0}0.60 &\cellcolor[HTML]{f7d7d4}0.56 &\cellcolor[HTML]{f6d1ce}0.54 &\cellcolor[HTML]{f1bab5}0.46 &\cellcolor[HTML]{f4c6c2}0.50 &\cellcolor[HTML]{f0b4af}0.44 &\cellcolor[HTML]{f9e2e0}0.60 &\cellcolor[HTML]{efafa9}0.42 &\cellcolor[HTML]{efafa9}0.42 \\
&3 &Dep &Int &\cellcolor[HTML]{84ceaa}0.92 &\cellcolor[HTML]{6ec59a}0.96 &\cellcolor[HTML]{8fd2b1}0.90 &\cellcolor[HTML]{8fd2b1}0.90 &\cellcolor[HTML]{b1e0c9}0.84 &\cellcolor[HTML]{a6dbc1}0.86 &\cellcolor[HTML]{c7e9d8}0.80 &\cellcolor[HTML]{c7e9d8}0.80 &\cellcolor[HTML]{fbedec}0.64 &\cellcolor[HTML]{fbedec}0.64 &0.70 \\
&5 &Ran &Swap &\cellcolor[HTML]{8fd2b1}0.90 &\cellcolor[HTML]{9bd7b9}0.88 &\cellcolor[HTML]{fcf0ef}0.65 &\cellcolor[HTML]{f9dfdd}0.59 &\cellcolor[HTML]{f6d1ce}0.54 &\cellcolor[HTML]{f4c6c2}0.50 &\cellcolor[HTML]{f5cbc8}0.52 &\cellcolor[HTML]{f6d1ce}0.54 &\cellcolor[HTML]{f0b4af}0.44 &\cellcolor[HTML]{f4c6c2}0.50 &\cellcolor[HTML]{f4c6c2}0.50 \\
&5 &Dep &Swap &\cellcolor[HTML]{79c9a2}0.94 &\cellcolor[HTML]{bce4d1}0.82 &\cellcolor[HTML]{fae8e6}0.62 &\cellcolor[HTML]{fbedec}0.64 &\cellcolor[HTML]{f4c6c2}0.50 &\cellcolor[HTML]{f4c6c2}0.50 &\cellcolor[HTML]{f6d1ce}0.54 &\cellcolor[HTML]{f3c0bc}0.48 &\cellcolor[HTML]{f1bab5}0.46 &\cellcolor[HTML]{f5cbc8}0.52 &\cellcolor[HTML]{f4c6c2}0.50 \\
&5 &Dep &Int &\cellcolor[HTML]{63c092}0.98 &\cellcolor[HTML]{63c092}0.98 &\cellcolor[HTML]{c7e9d8}0.80 &\cellcolor[HTML]{a6dbc1}0.86 &\cellcolor[HTML]{9bd7b9}0.88 &\cellcolor[HTML]{f4fbf8}0.72 &\cellcolor[HTML]{def2e8}0.76 &\cellcolor[HTML]{f9e2e0}0.60 &\cellcolor[HTML]{f8dcda}0.58 &\cellcolor[HTML]{fae8e6}0.62 &\cellcolor[HTML]{fcf3f2}0.66 \\
&5 &Dep &L-Int &\cellcolor[HTML]{63c092}0.98 &\cellcolor[HTML]{8fd2b1}0.90 &\cellcolor[HTML]{9bd7b9}0.88 &\cellcolor[HTML]{bce4d1}0.82 &\cellcolor[HTML]{a6dbc1}0.86 &\cellcolor[HTML]{c7e9d8}0.80 &\cellcolor[HTML]{d3ede0}0.78 &\cellcolor[HTML]{d3ede0}0.78 &\cellcolor[HTML]{f4fbf8}0.72 &0.70 &\cellcolor[HTML]{f4fbf8}0.72 \\

\midrule \midrule

\multirow{7}{*}{\begin{sideways} GPT 4o \end{sideways}} &3 &Ran &Swap &\cellcolor[HTML]{8fd2b1}0.90 &\cellcolor[HTML]{57bb8a}1.00 &\cellcolor[HTML]{57bb8a}1.00 &\cellcolor[HTML]{57bb8a}1.00 &\cellcolor[HTML]{6ec59a}0.96 &\cellcolor[HTML]{57bb8a}1.00 &\cellcolor[HTML]{57bb8a}1.00 &\cellcolor[HTML]{57bb8a}1.00 &\cellcolor[HTML]{57bb8a}1.00 &\cellcolor[HTML]{63c092}0.98 &\cellcolor[HTML]{57bb8a}1.00 \\
&3 &Dep &Swap &\cellcolor[HTML]{57bb8a}1.00 &\cellcolor[HTML]{84ceaa}0.92 &\cellcolor[HTML]{63c092}0.98 &\cellcolor[HTML]{57bb8a}1.00 &\cellcolor[HTML]{57bb8a}1.00 &\cellcolor[HTML]{57bb8a}1.00 &\cellcolor[HTML]{57bb8a}1.00 &\cellcolor[HTML]{57bb8a}1.00 &\cellcolor[HTML]{57bb8a}1.00 &\cellcolor[HTML]{63c092}0.98 &\cellcolor[HTML]{57bb8a}1.00 \\
&3 &Dep &Int &\cellcolor[HTML]{57bb8a}1.00 &\cellcolor[HTML]{57bb8a}1.00 &\cellcolor[HTML]{57bb8a}1.00 &\cellcolor[HTML]{57bb8a}1.00 &\cellcolor[HTML]{57bb8a}1.00 &\cellcolor[HTML]{57bb8a}1.00 &\cellcolor[HTML]{57bb8a}1.00 &\cellcolor[HTML]{57bb8a}1.00 &\cellcolor[HTML]{57bb8a}1.00 &\cellcolor[HTML]{57bb8a}1.00 &\cellcolor[HTML]{57bb8a}1.00 \\
&5 &Ran &Swap &\cellcolor[HTML]{57bb8a}1.00 &\cellcolor[HTML]{63c092}0.98 &\cellcolor[HTML]{57bb8a}1.00 &\cellcolor[HTML]{6ec59a}0.96 &\cellcolor[HTML]{57bb8a}1.00 &\cellcolor[HTML]{57bb8a}1.00 &\cellcolor[HTML]{57bb8a}1.00 &\cellcolor[HTML]{57bb8a}1.00 &\cellcolor[HTML]{63c092}0.98 &\cellcolor[HTML]{57bb8a}1.00 &\cellcolor[HTML]{79c9a2}0.94 \\
&5 &Dep &Swap &\cellcolor[HTML]{57bb8a}1.00 &\cellcolor[HTML]{63c092}0.98 &\cellcolor[HTML]{57bb8a}1.00 &\cellcolor[HTML]{57bb8a}1.00 &\cellcolor[HTML]{57bb8a}1.00 &\cellcolor[HTML]{57bb8a}1.00 &\cellcolor[HTML]{6ec59a}0.96 &\cellcolor[HTML]{57bb8a}1.00 &\cellcolor[HTML]{57bb8a}1.00 &\cellcolor[HTML]{57bb8a}1.00 &\cellcolor[HTML]{6ec59a}0.96 \\
&5 &Dep &Int &\cellcolor[HTML]{57bb8a}1.00 &\cellcolor[HTML]{57bb8a}1.00 &\cellcolor[HTML]{57bb8a}1.00 &\cellcolor[HTML]{57bb8a}1.00 &\cellcolor[HTML]{57bb8a}1.00 &\cellcolor[HTML]{57bb8a}1.00 &\cellcolor[HTML]{63c092}0.98 &\cellcolor[HTML]{57bb8a}1.00 &\cellcolor[HTML]{57bb8a}1.00 &\cellcolor[HTML]{57bb8a}1.00 &\cellcolor[HTML]{6ec59a}0.96 \\
&5 &Dep &L-Int &\cellcolor[HTML]{63c092}0.98 &\cellcolor[HTML]{57bb8a}1.00 &\cellcolor[HTML]{57bb8a}1.00 &\cellcolor[HTML]{6ec59a}0.96 &\cellcolor[HTML]{57bb8a}1.00 &\cellcolor[HTML]{63c092}0.98 &\cellcolor[HTML]{57bb8a}1.00 &\cellcolor[HTML]{57bb8a}1.00 &\cellcolor[HTML]{63c092}0.98 &\cellcolor[HTML]{57bb8a}1.00 &\cellcolor[HTML]{57bb8a}1.00 \\

\midrule \midrule

\multirow{7}{*}{\begin{sideways} \bf GPT 4o CoT \end{sideways}} &3 &Ran &Swap &\cellcolor[HTML]{57bb8a}1.00 &\cellcolor[HTML]{63c092}0.98 &\cellcolor[HTML]{57bb8a}1.00 &\cellcolor[HTML]{57bb8a}1.00 &\cellcolor[HTML]{6ec59a}0.96 &\cellcolor[HTML]{57bb8a}1.00 &\cellcolor[HTML]{57bb8a}1.00 &\cellcolor[HTML]{79c9a2}0.94 &\cellcolor[HTML]{57bb8a}1.00 &\cellcolor[HTML]{6ec59a}0.96 &\cellcolor[HTML]{63c092}0.98 \\
&3 &Dep &Swap &\cellcolor[HTML]{57bb8a}1.00 &\cellcolor[HTML]{57bb8a}1.00 &\cellcolor[HTML]{63c092}0.98 &\cellcolor[HTML]{57bb8a}1.00 &\cellcolor[HTML]{57bb8a}1.00 &\cellcolor[HTML]{6ec59a}0.96 &\cellcolor[HTML]{6ec59a}0.96 &\cellcolor[HTML]{57bb8a}1.00 &\cellcolor[HTML]{79c9a2}0.94 &\cellcolor[HTML]{57bb8a}1.00 &\cellcolor[HTML]{57bb8a}1.00 \\
&3 &Dep &Int &\cellcolor[HTML]{57bb8a}1.00 &\cellcolor[HTML]{57bb8a}1.00 &\cellcolor[HTML]{57bb8a}1.00 &\cellcolor[HTML]{57bb8a}1.00 &\cellcolor[HTML]{57bb8a}1.00 &\cellcolor[HTML]{57bb8a}1.00 &\cellcolor[HTML]{63c092}0.98 &\cellcolor[HTML]{57bb8a}1.00 &\cellcolor[HTML]{57bb8a}1.00 &\cellcolor[HTML]{6ec59a}0.96 &\cellcolor[HTML]{57bb8a}1.00 \\
&5 &Ran &Swap &\cellcolor[HTML]{63c092}0.98 &\cellcolor[HTML]{57bb8a}1.00 &\cellcolor[HTML]{57bb8a}1.00 &\cellcolor[HTML]{63c092}0.98 &\cellcolor[HTML]{57bb8a}1.00 &\cellcolor[HTML]{57bb8a}1.00 &\cellcolor[HTML]{57bb8a}1.00 &\cellcolor[HTML]{57bb8a}1.00 &\cellcolor[HTML]{57bb8a}1.00 &\cellcolor[HTML]{57bb8a}1.00 &\cellcolor[HTML]{57bb8a}1.00 \\
&5 &Dep &Swap &\cellcolor[HTML]{57bb8a}1.00 &\cellcolor[HTML]{57bb8a}1.00 &\cellcolor[HTML]{63c092}0.98 &\cellcolor[HTML]{57bb8a}1.00 &\cellcolor[HTML]{6ec59a}0.96 &\cellcolor[HTML]{57bb8a}1.00 &\cellcolor[HTML]{57bb8a}1.00 &\cellcolor[HTML]{57bb8a}1.00 &\cellcolor[HTML]{63c092}0.98 &\cellcolor[HTML]{57bb8a}1.00 &\cellcolor[HTML]{6ec59a}0.96 \\
&5 &Dep &Int &\cellcolor[HTML]{57bb8a}1.00 &\cellcolor[HTML]{57bb8a}1.00 &\cellcolor[HTML]{57bb8a}1.00 &\cellcolor[HTML]{57bb8a}1.00 &\cellcolor[HTML]{57bb8a}1.00 &\cellcolor[HTML]{57bb8a}1.00 &\cellcolor[HTML]{57bb8a}1.00 &\cellcolor[HTML]{57bb8a}1.00 &\cellcolor[HTML]{57bb8a}1.00 &\cellcolor[HTML]{57bb8a}1.00 &\cellcolor[HTML]{63c092}0.98 \\
&5 &Dep &L-Int &\cellcolor[HTML]{63c092}0.98 &\cellcolor[HTML]{57bb8a}1.00 &\cellcolor[HTML]{57bb8a}1.00 &\cellcolor[HTML]{79c9a2}0.94 &\cellcolor[HTML]{57bb8a}1.00 &\cellcolor[HTML]{57bb8a}1.00 &\cellcolor[HTML]{63c092}0.98 &\cellcolor[HTML]{57bb8a}1.00 &\cellcolor[HTML]{57bb8a}1.00 &\cellcolor[HTML]{6ec59a}0.96 &\cellcolor[HTML]{57bb8a}1.00 \\
\bottomrule
\end{tabular}}
    \caption{Accuracy at various depths of GPT models on \texttt{LinearWorld} task variations. Models evaluated with Chain-of-Thought prompting are denoted by "CoT" in their names. For the query types, "Ran" refers to "random" and "Dep" to "dependent" query type. Also, regarding the update types, "Int" and "L-Int" refer to "integer" and "large integer" displacement update types, respectively.}
    \label{tab:complementary_gpt}
\end{table*}

\begin{table*}[htb]
    \resizebox{\linewidth}{!}{\begin{tabular}{llll cccccccccccc}\toprule
\multirow{2}{*}{System} &\multirow{2}{*}{\# } &\multirow{2}{*}{Query} &\multirow{2}{*}{Update} &\multicolumn{11}{c}{Depth} \\ \cmidrule(rl){5-15} 

& & & &0 &1 &2 &3 &4 &5 &6 &7 &8 &9 &10 \\ 

\midrule

\multirow{7}{*}{\begin{sideways} Mixtral \end{sideways}} &3 &Ran &Swap &\cellcolor[HTML]{c7e9d8}0.80 &\cellcolor[HTML]{fdf9f8}0.68 &\cellcolor[HTML]{f3c0bc}0.48 &\cellcolor[HTML]{f3c0bc}0.48 &\cellcolor[HTML]{f8dcda}0.58 &\cellcolor[HTML]{f0b4af}0.44 &\cellcolor[HTML]{f1bab5}0.46 &\cellcolor[HTML]{efafa9}0.42 &\cellcolor[HTML]{f7d7d4}0.56 &\cellcolor[HTML]{f0b4af}0.44 &\cellcolor[HTML]{eea9a3}0.40 \\
&3 &Dep &Swap &\cellcolor[HTML]{f8dcda}0.58 &\cellcolor[HTML]{f8dcda}0.58 &\cellcolor[HTML]{f6d1ce}0.54 &\cellcolor[HTML]{f4c6c2}0.50 &\cellcolor[HTML]{f4c6c2}0.50 &\cellcolor[HTML]{efafa9}0.42 &\cellcolor[HTML]{f1bab5}0.46 &\cellcolor[HTML]{f3c0bc}0.48 &\cellcolor[HTML]{f1bab5}0.46 &\cellcolor[HTML]{f8dcda}0.58 &\cellcolor[HTML]{f9e2e0}0.60 \\
&3 &Dep &Int &\cellcolor[HTML]{f0b4af}0.44 &\cellcolor[HTML]{d3ede0}0.78 &\cellcolor[HTML]{fdf9f8}0.68 &\cellcolor[HTML]{fbedec}0.64 &\cellcolor[HTML]{f6d1ce}0.54 &\cellcolor[HTML]{f1bab5}0.46 &\cellcolor[HTML]{f0b4af}0.44 &\cellcolor[HTML]{f0b4af}0.44 &\cellcolor[HTML]{eda39d}0.38 &\cellcolor[HTML]{e67c73}0.24 &\cellcolor[HTML]{e8877f}0.28 \\
&5 &Ran &Swap &\cellcolor[HTML]{c7e9d8}0.80 &\cellcolor[HTML]{fbedec}0.64 &\cellcolor[HTML]{fbedec}0.64 &\cellcolor[HTML]{f8dcda}0.58 &\cellcolor[HTML]{f3c0bc}0.48 &\cellcolor[HTML]{f5cbc8}0.52 &\cellcolor[HTML]{f4c6c2}0.50 &\cellcolor[HTML]{f7d7d4}0.56 &\cellcolor[HTML]{f0b4af}0.44 &\cellcolor[HTML]{f5cbc8}0.52 &\cellcolor[HTML]{f7d7d4}0.56 \\
&5 &Dep &Swap &\cellcolor[HTML]{fdf9f8}0.68 &\cellcolor[HTML]{f4c6c2}0.50 &\cellcolor[HTML]{f4c6c2}0.50 &\cellcolor[HTML]{f4c6c2}0.50 &\cellcolor[HTML]{f8dcda}0.58 &\cellcolor[HTML]{eea9a3}0.40 &\cellcolor[HTML]{f8dcda}0.58 &\cellcolor[HTML]{f3c0bc}0.48 &\cellcolor[HTML]{eda39d}0.38 &\cellcolor[HTML]{f1bab5}0.46 &\cellcolor[HTML]{f4c6c2}0.50 \\
&5 &Dep &Int &\cellcolor[HTML]{fae8e6}0.62 &\cellcolor[HTML]{e9f6f0}0.74 &\cellcolor[HTML]{e9f6f0}0.74 &\cellcolor[HTML]{efafa9}0.42 &\cellcolor[HTML]{f6d1ce}0.54 &\cellcolor[HTML]{eea9a3}0.40 &\cellcolor[HTML]{f0b4af}0.44 &\cellcolor[HTML]{eda39d}0.38 &\cellcolor[HTML]{eda39d}0.38 &\cellcolor[HTML]{f1bab5}0.46 &\cellcolor[HTML]{ea928b}0.32 \\
&5 &Dep &L-Int &\cellcolor[HTML]{f4fbf8}0.72 &\cellcolor[HTML]{e9f6f0}0.74 &\cellcolor[HTML]{fcf3f2}0.66 &\cellcolor[HTML]{f0b4af}0.44 &\cellcolor[HTML]{fbedec}0.64 &\cellcolor[HTML]{f4c6c2}0.50 &\cellcolor[HTML]{efafa9}0.42 &\cellcolor[HTML]{f4c6c2}0.50 &\cellcolor[HTML]{fcf3f2}0.66 &\cellcolor[HTML]{f9e2e0}0.60 &\cellcolor[HTML]{f7d7d4}0.56 \\

\midrule \midrule

\multirow{7}{*}{\begin{sideways} Mixtral CoT \end{sideways}} &3 &Ran &Swap &\cellcolor[HTML]{8fd2b1}0.90 &\cellcolor[HTML]{bce4d1}0.82 &\cellcolor[HTML]{f6d1ce}0.54 &\cellcolor[HTML]{f6d1ce}0.54 &\cellcolor[HTML]{f3c0bc}0.48 &\cellcolor[HTML]{f8dcda}0.58 &\cellcolor[HTML]{efafa9}0.42 &\cellcolor[HTML]{eea9a3}0.40 &\cellcolor[HTML]{f6d1ce}0.54 &\cellcolor[HTML]{f3c0bc}0.48 &\cellcolor[HTML]{efafa9}0.42 \\
&3 &Dep &Swap &\cellcolor[HTML]{57bb8a}1.00 &\cellcolor[HTML]{e9f6f0}0.74 &\cellcolor[HTML]{f9e2e0}0.60 &\cellcolor[HTML]{fdf9f8}0.68 &\cellcolor[HTML]{f5cbc8}0.52 &\cellcolor[HTML]{f6d1ce}0.54 &\cellcolor[HTML]{f9e2e0}0.60 &\cellcolor[HTML]{ec9e97}0.36 &\cellcolor[HTML]{f8dcda}0.58 &\cellcolor[HTML]{efafa9}0.42 &\cellcolor[HTML]{fae8e6}0.62 \\
&3 &Dep &Int &\cellcolor[HTML]{fdf9f8}0.68 &\cellcolor[HTML]{d3ede0}0.78 &\cellcolor[HTML]{84ceaa}0.92 &\cellcolor[HTML]{bce4d1}0.82 &\cellcolor[HTML]{a6dbc1}0.86 &\cellcolor[HTML]{a6dbc1}0.86 &\cellcolor[HTML]{bce4d1}0.82 &\cellcolor[HTML]{d3ede0}0.78 &\cellcolor[HTML]{b1e0c9}0.84 &\cellcolor[HTML]{bce4d1}0.82 &\cellcolor[HTML]{d3ede0}0.78 \\
&5 &Ran &Swap &\cellcolor[HTML]{84ceaa}0.92 &\cellcolor[HTML]{a6dbc1}0.86 &\cellcolor[HTML]{f4fbf8}0.72 &\cellcolor[HTML]{f6d1ce}0.54 &\cellcolor[HTML]{f7d7d4}0.56 &\cellcolor[HTML]{f3c0bc}0.48 &\cellcolor[HTML]{f7d7d4}0.56 &\cellcolor[HTML]{f4c6c2}0.50 &\cellcolor[HTML]{f6d1ce}0.54 &\cellcolor[HTML]{f9e2e0}0.60 &\cellcolor[HTML]{f4c6c2}0.50 \\
&5 &Dep &Swap &\cellcolor[HTML]{63c092}0.98 &\cellcolor[HTML]{fcf3f2}0.66 &\cellcolor[HTML]{f6d1ce}0.54 &\cellcolor[HTML]{f5cbc8}0.52 &\cellcolor[HTML]{f7d7d4}0.56 &\cellcolor[HTML]{eda39d}0.38 &\cellcolor[HTML]{f4c6c2}0.50 &\cellcolor[HTML]{f4c6c2}0.50 &\cellcolor[HTML]{f9e2e0}0.60 &\cellcolor[HTML]{f6d1ce}0.54 &\cellcolor[HTML]{f6d1ce}0.54 \\
&5 &Dep &Int &\cellcolor[HTML]{c7e9d8}0.80 &\cellcolor[HTML]{bce4d1}0.82 &\cellcolor[HTML]{bce4d1}0.82 &\cellcolor[HTML]{9bd7b9}0.88 &\cellcolor[HTML]{9bd7b9}0.88 &\cellcolor[HTML]{9bd7b9}0.88 &\cellcolor[HTML]{bce4d1}0.82 &\cellcolor[HTML]{f4fbf8}0.72 &\cellcolor[HTML]{f4fbf8}0.72 &\cellcolor[HTML]{bce4d1}0.82 &\cellcolor[HTML]{def2e8}0.76 \\
&5 &Dep &L-Int &\cellcolor[HTML]{8fd2b1}0.90 &\cellcolor[HTML]{c7e9d8}0.80 &0.70 &\cellcolor[HTML]{d3ede0}0.78 &\cellcolor[HTML]{c7e9d8}0.80 &\cellcolor[HTML]{d3ede0}0.78 &\cellcolor[HTML]{fdf9f8}0.68 &\cellcolor[HTML]{f9e2e0}0.60 &\cellcolor[HTML]{fdf9f8}0.68 &\cellcolor[HTML]{f4fbf8}0.72 &\cellcolor[HTML]{c7e9d8}0.80 \\

\midrule \midrule

\multirow{7}{*}{\begin{sideways} Llama3 8B \end{sideways}} &3 &Ran &Swap &\cellcolor[HTML]{f1bab5}0.46 &\cellcolor[HTML]{f4c6c2}0.50 &\cellcolor[HTML]{f4c6c2}0.50 &\cellcolor[HTML]{efafa9}0.42 &\cellcolor[HTML]{f4c6c2}0.50 &\cellcolor[HTML]{f6d1ce}0.54 &\cellcolor[HTML]{f3c0bc}0.48 &\cellcolor[HTML]{f1bab5}0.46 &\cellcolor[HTML]{f5cbc8}0.52 &\cellcolor[HTML]{f7d7d4}0.56 &\cellcolor[HTML]{f0b4af}0.44 \\
&3 &Dep &Swap &\cellcolor[HTML]{f4c6c2}0.50 &\cellcolor[HTML]{f7d7d4}0.56 &\cellcolor[HTML]{f4c6c2}0.50 &\cellcolor[HTML]{f6d1ce}0.54 &\cellcolor[HTML]{f3c0bc}0.48 &\cellcolor[HTML]{f5cbc8}0.52 &\cellcolor[HTML]{f1bab5}0.46 &\cellcolor[HTML]{f1bab5}0.46 &\cellcolor[HTML]{efafa9}0.42 &\cellcolor[HTML]{f0b4af}0.44 &\cellcolor[HTML]{f5cbc8}0.52 \\
&3 &Dep &Int &\cellcolor[HTML]{f9e2e0}0.60 &\cellcolor[HTML]{f6d1ce}0.54 &\cellcolor[HTML]{fdf9f8}0.68 &\cellcolor[HTML]{c7e9d8}0.80 &\cellcolor[HTML]{f4fbf8}0.72 &\cellcolor[HTML]{e9f6f0}0.74 &0.70 &\cellcolor[HTML]{fbedec}0.64 &\cellcolor[HTML]{f4fbf8}0.72 &\cellcolor[HTML]{bce4d1}0.82 &\cellcolor[HTML]{fae8e6}0.62 \\
&5 &Ran &Swap &\cellcolor[HTML]{f4c6c2}0.50 &\cellcolor[HTML]{f7d7d4}0.56 &\cellcolor[HTML]{f6d1ce}0.54 &\cellcolor[HTML]{f6d1ce}0.54 &\cellcolor[HTML]{ec9e97}0.36 &\cellcolor[HTML]{eda39d}0.38 &\cellcolor[HTML]{f7d7d4}0.56 &\cellcolor[HTML]{f3c0bc}0.48 &\cellcolor[HTML]{f5cbc8}0.52 &\cellcolor[HTML]{f3c0bc}0.48 &\cellcolor[HTML]{f5cbc8}0.52 \\
&5 &Dep &Swap &\cellcolor[HTML]{f4c6c2}0.50 &\cellcolor[HTML]{f4c6c2}0.50 &\cellcolor[HTML]{f5cbc8}0.52 &\cellcolor[HTML]{f6d1ce}0.54 &\cellcolor[HTML]{f4c6c2}0.50 &\cellcolor[HTML]{f0b4af}0.44 &\cellcolor[HTML]{f4c6c2}0.50 &\cellcolor[HTML]{f4c6c2}0.50 &\cellcolor[HTML]{f6d1ce}0.54 &\cellcolor[HTML]{f7d7d4}0.56 &\cellcolor[HTML]{f8dcda}0.58 \\
&5 &Dep &Int &\cellcolor[HTML]{eda39d}0.38 &\cellcolor[HTML]{d3ede0}0.78 &\cellcolor[HTML]{f6d1ce}0.54 &\cellcolor[HTML]{f4fbf8}0.72 &\cellcolor[HTML]{f5cbc8}0.52 &\cellcolor[HTML]{fdf9f8}0.68 &\cellcolor[HTML]{f9e2e0}0.60 &\cellcolor[HTML]{fae8e6}0.62 &\cellcolor[HTML]{f7d7d4}0.56 &\cellcolor[HTML]{f6d1ce}0.54 &\cellcolor[HTML]{f9e2e0}0.60 \\
&5 &Dep &L-Int &\cellcolor[HTML]{fdf9f8}0.68 &\cellcolor[HTML]{f8dcda}0.58 &\cellcolor[HTML]{f5cbc8}0.52 &\cellcolor[HTML]{f4c6c2}0.50 &\cellcolor[HTML]{fae8e6}0.62 &\cellcolor[HTML]{f4c6c2}0.50 &\cellcolor[HTML]{f8dcda}0.58 &\cellcolor[HTML]{f7d7d4}0.56 &\cellcolor[HTML]{f5cbc8}0.52 &\cellcolor[HTML]{f6d1ce}0.54 &\cellcolor[HTML]{f6d1ce}0.54 \\

\midrule \midrule

\multirow{7}{*}{\begin{sideways} Llama3 8B CoT \end{sideways}} &3 &Ran &Swap &0.70 &\cellcolor[HTML]{f5cbc8}0.52 &\cellcolor[HTML]{f8dcda}0.58 &\cellcolor[HTML]{f0b4af}0.44 &\cellcolor[HTML]{f5cbc8}0.52 &\cellcolor[HTML]{f7d7d4}0.56 &\cellcolor[HTML]{f6d1ce}0.54 &\cellcolor[HTML]{fbedec}0.64 &\cellcolor[HTML]{f5cbc8}0.52 &\cellcolor[HTML]{efafa9}0.42 &\cellcolor[HTML]{f5cbc8}0.52 \\
&3 &Dep &Swap &\cellcolor[HTML]{f7d7d4}0.56 &\cellcolor[HTML]{f4c6c2}0.50 &\cellcolor[HTML]{fae8e6}0.62 &\cellcolor[HTML]{f5cbc8}0.52 &\cellcolor[HTML]{f6d1ce}0.54 &\cellcolor[HTML]{eda39d}0.38 &\cellcolor[HTML]{f8dcda}0.58 &\cellcolor[HTML]{f0b4af}0.44 &\cellcolor[HTML]{f7d7d4}0.56 &\cellcolor[HTML]{ea928b}0.32 &\cellcolor[HTML]{f3c0bc}0.48 \\
&3 &Dep &Int &\cellcolor[HTML]{ec9e97}0.36 &\cellcolor[HTML]{bce4d1}0.82 &\cellcolor[HTML]{c7e9d8}0.80 &\cellcolor[HTML]{def2e8}0.76 &\cellcolor[HTML]{fcf3f2}0.66 &\cellcolor[HTML]{fbedec}0.64 &0.70 &0.70 &\cellcolor[HTML]{e9f6f0}0.74 &0.70 &\cellcolor[HTML]{fae8e6}0.62 \\
&5 &Ran &Swap &\cellcolor[HTML]{f8dcda}0.58 &\cellcolor[HTML]{f0b4af}0.44 &\cellcolor[HTML]{fae8e6}0.62 &\cellcolor[HTML]{f1bab5}0.46 &\cellcolor[HTML]{f3c0bc}0.48 &\cellcolor[HTML]{f7d7d4}0.56 &\cellcolor[HTML]{f1bab5}0.46 &\cellcolor[HTML]{f7d7d4}0.56 &\cellcolor[HTML]{f4c6c2}0.50 &\cellcolor[HTML]{f3c0bc}0.48 &\cellcolor[HTML]{f7d7d4}0.56 \\
&5 &Dep &Swap &\cellcolor[HTML]{fcf3f2}0.66 &\cellcolor[HTML]{fae8e6}0.62 &\cellcolor[HTML]{f4c6c2}0.50 &\cellcolor[HTML]{fbedec}0.64 &\cellcolor[HTML]{f4c6c2}0.50 &\cellcolor[HTML]{f1bab5}0.46 &\cellcolor[HTML]{f8dcda}0.58 &\cellcolor[HTML]{f0b4af}0.44 &\cellcolor[HTML]{f3c0bc}0.48 &\cellcolor[HTML]{eea9a3}0.40 &\cellcolor[HTML]{f5cbc8}0.52 \\
&5 &Dep &Int &\cellcolor[HTML]{f1bab5}0.46 &\cellcolor[HTML]{e9f6f0}0.74 &\cellcolor[HTML]{d3ede0}0.78 &\cellcolor[HTML]{fcf3f2}0.66 &\cellcolor[HTML]{f8dcda}0.58 &0.70 &\cellcolor[HTML]{f5cbc8}0.52 &\cellcolor[HTML]{f8dcda}0.58 &\cellcolor[HTML]{f9e2e0}0.60 &\cellcolor[HTML]{f8dcda}0.58 &\cellcolor[HTML]{f3c0bc}0.48 \\
&5 &Dep &L-Int &\cellcolor[HTML]{f1bab5}0.46 &\cellcolor[HTML]{fdf9f8}0.68 &\cellcolor[HTML]{f7d7d4}0.56 &\cellcolor[HTML]{fae8e6}0.62 &\cellcolor[HTML]{fae8e6}0.62 &\cellcolor[HTML]{f9e2e0}0.60 &\cellcolor[HTML]{f6d1ce}0.54 &\cellcolor[HTML]{f5cbc8}0.52 &\cellcolor[HTML]{f1bab5}0.46 &\cellcolor[HTML]{f1bab5}0.46 &\cellcolor[HTML]{efafa9}0.42 \\

\midrule \midrule

\multirow{7}{*}{\begin{sideways} Llama3 70B \end{sideways}} 	&3 &Ran &Swap &\cellcolor[HTML]{57bb8a}1 &\cellcolor[HTML]{bce4d1}0.82 &\cellcolor[HTML]{f6d1ce}0.54 &\cellcolor[HTML]{eb9891}0.34 &\cellcolor[HTML]{f7d7d4}0.56 &\cellcolor[HTML]{fcf3f2}0.66 &\cellcolor[HTML]{f5cbc8}0.52 &\cellcolor[HTML]{f5cbc8}0.52 &\cellcolor[HTML]{fae8e6}0.62 &\cellcolor[HTML]{f7d7d4}0.56 &\cellcolor[HTML]{f5cbc8}0.52 \\
&3 &Dep &Swap &\cellcolor[HTML]{57bb8a}1 &\cellcolor[HTML]{f3c0bc}0.48 &\cellcolor[HTML]{f3c0bc}0.48 &\cellcolor[HTML]{f4c6c2}0.5 &\cellcolor[HTML]{f8dcda}0.58 &\cellcolor[HTML]{fae8e6}0.62 &\cellcolor[HTML]{efafa9}0.42 &\cellcolor[HTML]{f6d1ce}0.54 &\cellcolor[HTML]{f0b4af}0.44 &\cellcolor[HTML]{f4c6c2}0.5 &\cellcolor[HTML]{f7d7d4}0.56 \\
&3 &Dep &Int &\cellcolor[HTML]{79c9a2}0.94 &\cellcolor[HTML]{57bb8a}1 &\cellcolor[HTML]{6ec59a}0.96 &\cellcolor[HTML]{6ec59a}0.96 &\cellcolor[HTML]{79c9a2}0.94 &\cellcolor[HTML]{6ec59a}0.96 &\cellcolor[HTML]{6ec59a}0.96 &\cellcolor[HTML]{63c092}0.98 &\cellcolor[HTML]{8fd2b1}0.9 &\cellcolor[HTML]{84ceaa}0.92 &\cellcolor[HTML]{9bd7b9}0.88 \\
&5 &Ran &Swap &\cellcolor[HTML]{63c092}0.98 &\cellcolor[HTML]{f4fbf8}0.72 &\cellcolor[HTML]{a6dbc1}0.86 &\cellcolor[HTML]{f6d1ce}0.54 &\cellcolor[HTML]{c7e9d8}0.8 &\cellcolor[HTML]{d3ede0}0.78 &\cellcolor[HTML]{f8dcda}0.58 &\cellcolor[HTML]{fcf3f2}0.66 &\cellcolor[HTML]{fcf3f2}0.66 &\cellcolor[HTML]{f8dcda}0.58 &0.7 \\
&5 &Dep &Swap &\cellcolor[HTML]{57bb8a}1 &\cellcolor[HTML]{f4c6c2}0.5 &\cellcolor[HTML]{f7d7d4}0.56 &\cellcolor[HTML]{fcf3f2}0.66 &\cellcolor[HTML]{fae8e6}0.62 &\cellcolor[HTML]{fae8e6}0.62 &\cellcolor[HTML]{f5cbc8}0.52 &\cellcolor[HTML]{f6d1ce}0.54 &\cellcolor[HTML]{f7d7d4}0.56 &\cellcolor[HTML]{f6d1ce}0.54 &\cellcolor[HTML]{f9e2e0}0.6 \\
&5 &Dep &Int &\cellcolor[HTML]{63c092}0.98 &\cellcolor[HTML]{79c9a2}0.94 &\cellcolor[HTML]{8fd2b1}0.9 &\cellcolor[HTML]{79c9a2}0.94 &\cellcolor[HTML]{6ec59a}0.96 &\cellcolor[HTML]{63c092}0.98 &\cellcolor[HTML]{84ceaa}0.92 &\cellcolor[HTML]{6ec59a}0.96 &\cellcolor[HTML]{6ec59a}0.96 &\cellcolor[HTML]{79c9a2}0.94 &\cellcolor[HTML]{63c092}0.98 \\
&5 &Dep &L-Int &\cellcolor[HTML]{8fd2b1}0.9 &\cellcolor[HTML]{79c9a2}0.94 &\cellcolor[HTML]{63c092}0.98 &\cellcolor[HTML]{6ec59a}0.96 &\cellcolor[HTML]{8fd2b1}0.9 &\cellcolor[HTML]{8fd2b1}0.9 &\cellcolor[HTML]{84ceaa}0.92 &\cellcolor[HTML]{6ec59a}0.96 &\cellcolor[HTML]{79c9a2}0.94 &\cellcolor[HTML]{9bd7b9}0.88 &\cellcolor[HTML]{bce4d1}0.82 \\

\midrule \midrule

\multirow{7}{*}{\begin{sideways} Llama3 70B CoT \end{sideways}} &3 &Ran &Swap &\cellcolor[HTML]{6ec59a}0.96 &\cellcolor[HTML]{a6dbc1}0.86 &\cellcolor[HTML]{9bd7b9}0.88 &\cellcolor[HTML]{c7e9d8}0.8 &\cellcolor[HTML]{d3ede0}0.78 &\cellcolor[HTML]{8fd2b1}0.9 &\cellcolor[HTML]{bce4d1}0.82 &\cellcolor[HTML]{a6dbc1}0.86 &\cellcolor[HTML]{def2e8}0.76 &\cellcolor[HTML]{f4fbf8}0.72 &\cellcolor[HTML]{b1e0c9}0.84 \\
&3 &Dep &Swap &\cellcolor[HTML]{57bb8a}1 &\cellcolor[HTML]{def2e8}0.76 &\cellcolor[HTML]{8fd2b1}0.9 &\cellcolor[HTML]{79c9a2}0.94 &\cellcolor[HTML]{8fd2b1}0.9 &\cellcolor[HTML]{9bd7b9}0.88 &\cellcolor[HTML]{79c9a2}0.94 &\cellcolor[HTML]{bce4d1}0.82 &\cellcolor[HTML]{def2e8}0.76 &\cellcolor[HTML]{d3ede0}0.78 &\cellcolor[HTML]{b1e0c9}0.84 \\
&3 &Dep &Int &\cellcolor[HTML]{79c9a2}0.94 &\cellcolor[HTML]{57bb8a}1 &\cellcolor[HTML]{57bb8a}1 &\cellcolor[HTML]{57bb8a}1 &\cellcolor[HTML]{6ec59a}0.96 &\cellcolor[HTML]{6ec59a}0.96 &\cellcolor[HTML]{8fd2b1}0.9 &\cellcolor[HTML]{6ec59a}0.96 &\cellcolor[HTML]{79c9a2}0.94 &\cellcolor[HTML]{6ec59a}0.96 &\cellcolor[HTML]{84ceaa}0.92 \\
&5 &Ran &Swap &\cellcolor[HTML]{6ec59a}0.96 &\cellcolor[HTML]{79c9a2}0.94 &\cellcolor[HTML]{63c092}0.98 &\cellcolor[HTML]{8fd2b1}0.9 &\cellcolor[HTML]{6ec59a}0.96 &\cellcolor[HTML]{8fd2b1}0.9 &\cellcolor[HTML]{9bd7b9}0.88 &\cellcolor[HTML]{63c092}0.98 &\cellcolor[HTML]{6ec59a}0.96 &\cellcolor[HTML]{6ec59a}0.96 &\cellcolor[HTML]{9bd7b9}0.88 \\
&5 &Dep &Swap &\cellcolor[HTML]{57bb8a}1 &\cellcolor[HTML]{b1e0c9}0.84 &\cellcolor[HTML]{57bb8a}1 &\cellcolor[HTML]{6ec59a}0.96 &\cellcolor[HTML]{84ceaa}0.92 &\cellcolor[HTML]{9bd7b9}0.88 &\cellcolor[HTML]{9bd7b9}0.88 &\cellcolor[HTML]{79c9a2}0.94 &\cellcolor[HTML]{9bd7b9}0.88 &\cellcolor[HTML]{c7e9d8}0.8 &\cellcolor[HTML]{6ec59a}0.96 \\
&5 &Dep &Int &\cellcolor[HTML]{63c092}0.98 &\cellcolor[HTML]{57bb8a}1 &\cellcolor[HTML]{57bb8a}1 &\cellcolor[HTML]{63c092}0.98 &\cellcolor[HTML]{57bb8a}1 &\cellcolor[HTML]{57bb8a}1 &\cellcolor[HTML]{57bb8a}1 &\cellcolor[HTML]{63c092}0.98 &\cellcolor[HTML]{6ec59a}0.96 &\cellcolor[HTML]{63c092}0.98 &\cellcolor[HTML]{79c9a2}0.94 \\
&5 &Dep &L-Int &\cellcolor[HTML]{6ec59a}0.96 &\cellcolor[HTML]{84ceaa}0.92 &\cellcolor[HTML]{8fd2b1}0.9 &\cellcolor[HTML]{84ceaa}0.92 &\cellcolor[HTML]{79c9a2}0.94 &\cellcolor[HTML]{6ec59a}0.96 &\cellcolor[HTML]{8fd2b1}0.9 &\cellcolor[HTML]{79c9a2}0.94 &\cellcolor[HTML]{9bd7b9}0.88 &\cellcolor[HTML]{b1e0c9}0.84 &\cellcolor[HTML]{84ceaa}0.92 \\

\bottomrule
\end{tabular}}
    \caption{Accuracy at various depths of open-source models on \texttt{LinearWorld} task variations. Models evaluated with Chain-of-Thought prompting are denoted by "CoT" in their names. For the query types, "Ran" refers to "random" and "Dep" to "dependent" query type. Also, regarding the update types, "Int" and "L-Int" refer to "integer" and "large integer" displacement update types, respectively.}
    \label{tab:complementary_open}
\end{table*}

\end{document}